\documentclass[10pt,twocolumn,letterpaper]{article}

\usepackage{iccv}              

\usepackage{adjustbox}
\usepackage{multirow}
\usepackage{url}
\usepackage{subcaption}
\usepackage{multirow}
\usepackage{tikz}
\usepackage{graphicx}
\usepackage{url}
\usepackage{booktabs}
\usepackage{tabularx}
\usepackage{wrapfig}
\usepackage{enumitem}
\usepackage{amsmath}
\usepackage{amsfonts}
\usepackage{bm}
\usepackage{graphicx}
\usepackage{enumerate}
\usepackage{caption}
\usepackage{graphicx}
\usepackage{multirow}
\usepackage{tabu}
\usepackage{xcolor}
\usepackage{booktabs}
\usepackage{colortbl}
\usepackage{algpseudocode}
\usepackage{algorithm}
\usepackage{footnote}
\usepackage{pifont}

\usepackage{float}

\definecolor{Red}{rgb}{0.6,0,0}
\definecolor{Blue}{rgb}{0,0,0.8}
\definecolor{Green}{rgb}{0,0.6,0.9}
\definecolor{airforceblue}{rgb}{0.36, 0.54, 0.66}
\definecolor{ao(english)}{rgb}{0.0, 0.5, 0.0}
\definecolor{azure(colorwheel)}{rgb}{0.0, 0.5, 1.0}
\definecolor{crimson}{rgb}{0.86, 0.08, 0.24}
\definecolor{darkcerulean}{rgb}{0.03, 0.27, 0.49}
\definecolor{cobalt}{rgb}{0.0, 0.28, 0.67}
\definecolor{rosegold}{rgb}{0.72, 0.43, 0.47}
\definecolor{orange-red}{rgb}{1.0, 0.27, 0.0}
\definecolor{mountainmeadow}{rgb}{0.19, 0.73, 0.56}
\definecolor{malachite}{rgb}{0.04, 0.85, 0.32}
\definecolor{darkblue}{rgb}{0.0, 0.0, 0.55}
\definecolor{customred}{rgb}{1, 0.85, 0.85}
\definecolor{customcitecolor}{rgb}{0.7, 0.5, 1}
\definecolor{custompink}{rgb}{0.8, 0.3, 0.3}
\definecolor{customgreen}{rgb}{0.3, 0.8, 0.3}
\definecolor{Lightgreen}{rgb}{0.8, 1, 0.9}
\definecolor{LightCyan}{rgb}{0.8, 0.9, 1}
\definecolor{mygray}{gray}{0.8}

\usepackage{nicematrix,tikz}
\definecolor{nicebluehsb}{HSB}{215,62,67}
\definecolor{nicebluergb}{RGB}{65,109,171}

\usepackage{mathrsfs}

\usepackage{nicefrac}       
\usepackage{booktabs}       

\usepackage{thmtools,thm-restate}

\usepackage{listings}
\usepackage{lstautogobble}  %
\usepackage{color}          %
\usepackage{zi4}            %

\usepackage{siunitx}               
\usepackage{url}
\usepackage{tikz, pgfplots}
\usetikzlibrary{positioning}
\usepackage{capt-of}
\usepackage{diagbox}

\declaretheorem[]{definition}

\def\eqref#1{(\ref{#1})}

\usepackage{colortbl}
\usepackage{mdframed}

\makeatletter
\renewcommand{\paragraph}{%
  \@startsection{paragraph}{4}%
  {\z@}{3.25ex \@plus 1ex \@minus .2ex}{-1.8em}%
  {\normalfont\normalsize\bfseries}%
}
\makeatother

\definecolor{iccvblue}{rgb}{0.21,0.49,0.74}
\usepackage[pagebackref,breaklinks,colorlinks,allcolors=iccvblue]{hyperref}

\def\paperID{8752} 
\def\confName{ICCV}
\def\confYear{2025}

\title{Forging and Removing Latent-Noise Diffusion Watermarks Using a Single Image}

\author{Anubhav Jain$^{1}$, Yuya Kobayashi$^{2}$, Naoki Murata$^{2}$, Yuhta Takida$^{2}$,  Takashi Shibuya$^{2}$,\\  Yuki Mitsufuji$^{2,3}$, Niv Cohen$^{1,*}$, Nasir Memon$^{1,*}$, Julian Togelius$^{1,}$\thanks{Equal Supervision} \\ 
$^1$New York University, $^2$Sony AI, $^3$Sony Group Corporation\\
\small \texttt{\{aj3281,nc3468,memon,julian.togelius\}@nyu.edu}\\ \small \texttt{\{u.kobayashi,naoki.murata,yuta.takida,takashi.tak.shibuya,yuhki.mitsufuji\}@sony.com}
}

\begin{document}
\maketitle
\begin{abstract}

Watermarking techniques are vital for protecting intellectual property and preventing fraudulent use of media.
Most previous watermarking schemes designed for diffusion models embed a secret key in the initial noise. The resulting pattern is often considered hard to remove and forge into unrelated images.
In this paper, we propose a black-box adversarial attack without presuming access to the diffusion model weights. Our attack uses only a single watermarked example and is based on a simple observation: there is a many-to-one mapping between images and initial noises. There are regions in the clean image latent space pertaining to each watermark that get mapped to the same initial noise when inverted. 
Based on this intuition, we propose an adversarial attack to forge the watermark by introducing perturbations to the images such that we can enter the region of watermarked images. We show that we can also apply a similar approach for watermark removal by learning perturbations to exit this region.   
We report results on multiple watermarking schemes (Tree-Ring, RingID, WIND, and Gaussian Shading) across two diffusion models (SDv1.4 and SDv2.0). Our results demonstrate the effectiveness of the attack and expose vulnerabilities in the watermarking methods, motivating future research on improving them. Our codebase is publicly available at \href{https://github.com/anubhav1997/watermark_forgery_removal}{https://github.com/anubhav1997/watermark\_forgery\_removal}. 

\end{abstract}


\begin{figure}
    \centering
    \begin{subfigure}[t]{\linewidth}
        \centering
        \includegraphics[trim={2.2cm 0.5cm 0.7cm 0cm},clip,width=\linewidth]{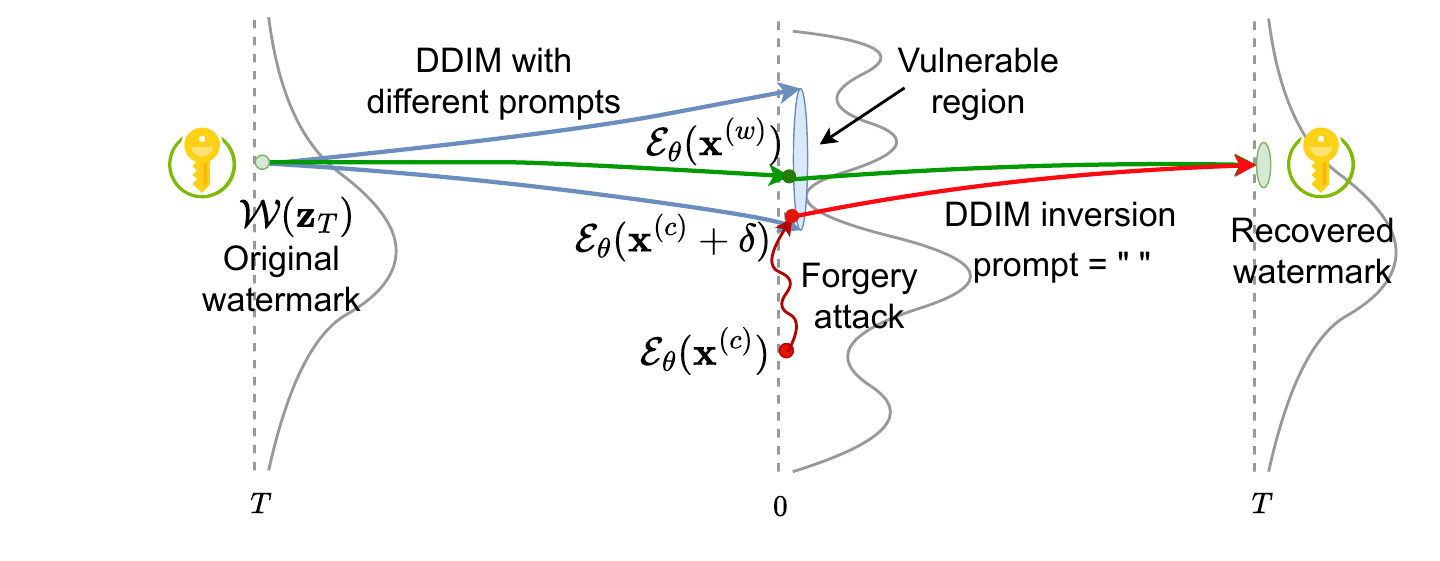}
    \end{subfigure}
    
    \caption{Intuition behind the attack: A whole region exists in the clean latent space, all of which approximately maps to a single key-embedded initial latent noise vector. A forgerer only needs to ensure their sample embeds within this region to be falsely classified as watermarked. }
    \label{fig:main_fig}
\end{figure}

\begin{figure*}
    \centering
    \includegraphics[width=0.7\linewidth]{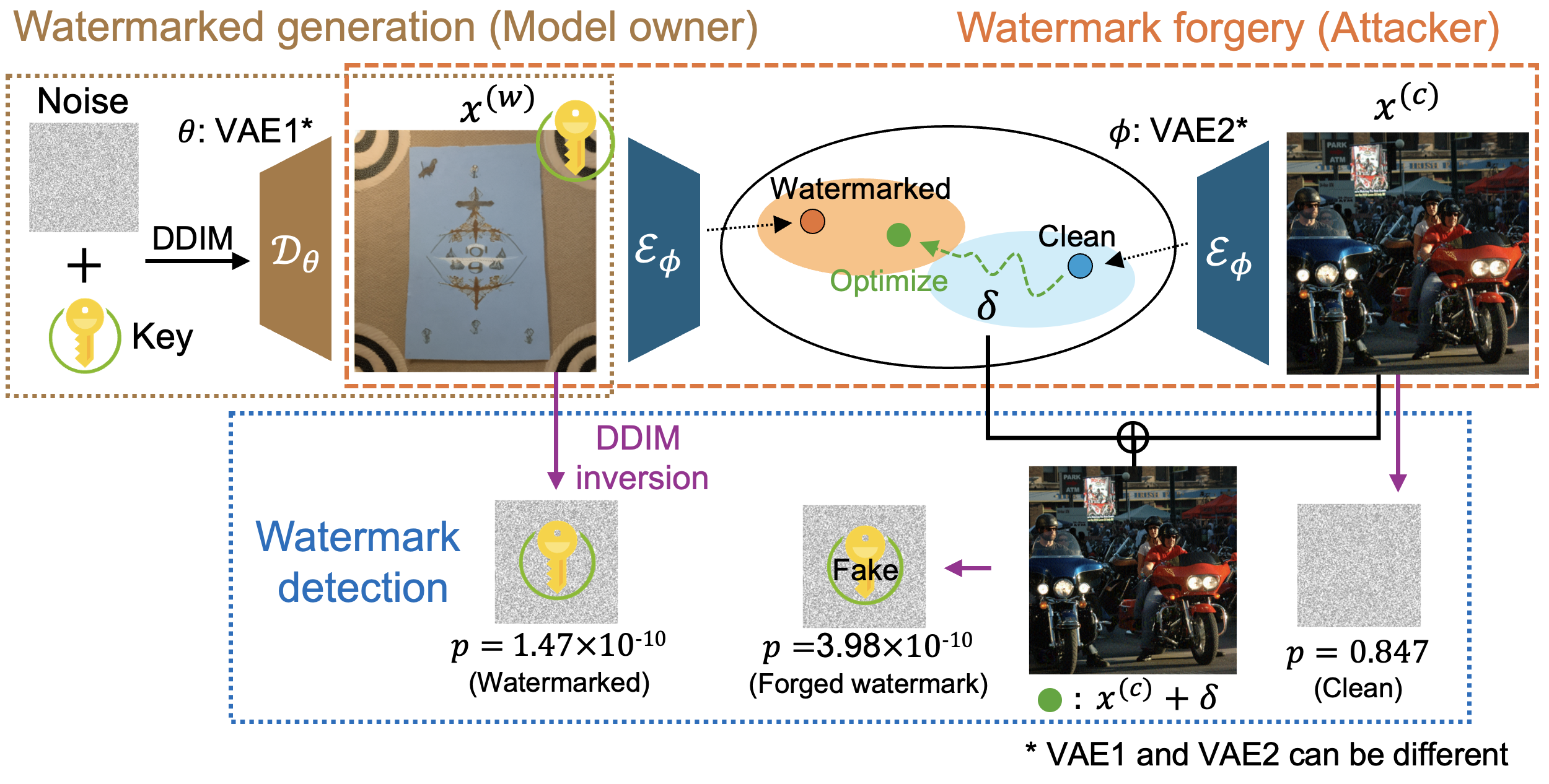}
    \caption{Our forgery attack works by finding an adversarial perturbation $\bm{\delta}$ such that the latent representation of the non-watermarked image $\mathcal{E}_{\phi}(\mathbf{x}^{(c)} + \bm{\delta})$ is close to the one corresponding to that of a watermarked image $\mathcal{E}_{\phi}(\mathbf{x}^{(w)})$. We do so while ensuring that we only introduce imperceptible changes to the clean image.  }
    \label{fig:overview}
\end{figure*}

\section{Introduction}

Latent diffusion models have made significant strides in terms of producing realistic-looking images. However, this advancement comes with its own set of problems, primarily related to image provenance. Image forensics experts are left with the task of verifying which generative model generated a particular image, if any, and who holds accountability for any harm arising out of it. Researchers and policymakers alike have focused on watermarking generative models as a potential solution to the threat from these models. This approach involves embedding an imperceptible pattern into images, allowing them to verify whether or not an image was generated using a specific model. 

Image watermarking is a well-studied field \cite{potdar2005survey,hartung1999multimedia,podilchuk2001digital}; however, recent advances in generative modeling have transformed both watermarking techniques and the attacks against them.
Leading watermarking schemes in diffusion models have focused on embedding a secret key into the initial noisy latent vector \cite{wen2023tree,ci2024ringid,arabi2024hidden,yang2024gaussian}. This allows the model owner to generate a realistic image using the standard denoising process without changing the model weights or compromising the image quality. During verification, the image is inverted to recover the initial noise sample and thus the initial pattern that might have been embedded. These methods have shown promising results as they avoid watermark removal by image transformations. Yet, as we show in this paper, the process still remains vulnerable to adversaries.

Attacks on watermarks fall into two main categories:  \textit{forgery attacks}~\citep{yang2024steganalysis,kaur2023image,wang2021watermark} and \textit{removal attacks} \citep{zhao2025invisible,lukas2023leveraging,liu2024image,yang2024steganalysis,hu2024transfer,anwaves}. Forgery attacks attempt to steal the watermark and apply it to images unrelated to the model's owner, raising concerns about false attribution of harmful content. Removal attacks seek to remove the watermark while preserving the image content, leading to concerns about intellectual property infringement or harmful use of the generated content such as deepfakes.
Previous attack methods have shown some success, but their success relied on one or more of the following conditions: (i) collecting a large set of watermarked images (and possibly, their non-watermarked counterparts), (ii) access to the entire model weights, or some approximation of it, and (iii) significantly distorting the attacked images.

In this paper, we suggest a new attack method to forge the watermark assuming access only to a single watermarked image. We also apply a similar approach to remove the watermark. Our attack utilizes the watermarking scheme's reliance on DDIM noise inversion \cite{mokady2023null}. Namely, we utilize the fact that different prompts can utilize the same initial noise during generation, but the DDIM inversion takes place with an empty prompt. This fact implies that there is a many-to-one relationship from the clean denoised latent space to the initial noise latent space. This in turn suggests that there might be a non-trivial latent region in a Variational Auto-encoder's (VAE) latent space that corresponds to each watermark. We showcase this idea in Figure \ref{fig:main_fig}. We verify that this hypothesis is true by using a linear support vector machine model to find latent directions corresponding to watermarked and non-watermarked images. Traversing along these directions allows easy removal and forgery of the watermark signal. 

Building upon this, we propose an adversarial attack to forge a given watermark, wherein the attacker simply needs to find adversarial perturbations such that an adversarial example gets embedded close enough to a watermarked example in the VAE's latent representation space. We show that an imperceptible modification to the image, which does not alter its semantic content, is sufficient due to the inherent non-smoothness of the VAE's representation space \cite{cemgil2020adversarially}. Once this objective is achieved, the DDIM inversion process with an empty prompt will guide both of these latents to a similar initial noise state, successfully fooling the detection system. We show a pictorial representation of our methodology in Figure \ref{fig:overview}. 

Moreover, we show that a similar method can be used for a removal attack, removing the watermark from an image while preserving the image content. In this scenario, our attack objective is to ensure that the latent representation for a watermarked image is as close to a non-watermarked image's latent as possible so that it leads to a false negative match when inverted. 

Thus, we show that our method only needs one watermarked image to forge the watermark and we can apply a similar idea to remove it. We can do so without fully inverting the image to its initial noise state to tamper with the secret key, which would require access to a denoising network used in generating the image. 
Although our method does require access to a VAE, it does not necessarily need access to the same VAE that was used by the watermarked diffusion model. A VAE that was trained on a similar dataset suffices for this task. Furthermore, our method does not assume any access to the denoising network (U-Net), which is typically fine-tuned for different applications such as avoiding generation of not-safe-for-work content and copyrighted images.




To summarize, we make the following contributions:
\begin{itemize}
    \item \textbf{Identifying a Watermarked Latent Subspaces} – We show that there exist latent directions and regions corresponding to each watermark pattern. 
    \item \textbf{Watermark Forgery and Removal Attacks} – We propose attacks that rely \textit{only} on a single watermarked image and a proxy VAE.
    \item \textbf{Impact Assessment} – We evaluate and minimize the impact of these attacks on the quality of the target image.
\end{itemize}

\section{Related Work}

\paragraph{Watermarking Schemes for Diffusion Models.}

Watermarking is a technique to ensure that an image can be traced back to its owner, helping with media authentication and preserving intellectual property \cite{potdar2005survey,hartung1999multimedia,podilchuk2001digital,wong2001secret,cox2007digital}. Classical watermarking techniques have involved embedding an invisible pattern in the image that can be recovered, and similarly, in the context of generative models, watermarks can be embedded post hoc after the image is generated \cite{wong2001secret,tancik2020stegastamp,cox2007digital,fernandez2022watermarking,zhang2019robust}. 
Watermarking schemes in diffusion models have also focused on directly or indirectly fine-tuning the VAE decoder using an adapter to watermark the output image \cite{ci2024wmadapter,fernandez2023stable,xiong2023flexible}. 


A popular line of work involves embedding a secret key into the initial noise during the denoising phase. Wen et al. \cite{wen2023tree}, who had initially proposed this idea, showed that initial noise-based watermarking is more robust against removal attacks - transformation aiming to render the watermark undetectable. Ci et al. \cite{ci2024ringid} further improved the watermarking pattern structure to make it more secure.  Yang et al. \cite{yang2024gaussian} utilized distribution-preserving sampling to ensure that the initial noise follows a Gaussian distribution, preserving the distribution of generated images. Arabi et al. \cite{arabi2024hidden} showed that breaking keys into groups and using an initial pattern specific to the groups allowed embedding a larger number of secret keys, improving security. Gunn et al. \cite{gunn2024undetectable} used a pseudo-random error correcting code to initialize the initial noise sample. These methods have focused on generating distortion-free images, coming from a similar distribution as non-watermarked images. It was often implicitly assumed that such distortion-free watermarks would be more secure against various attacks. However, as we will show, this might not be the case. 

In this paper, we focus on initial noise-based watermarking schemes that embed a secret key in the initial noise as they are often considered relatively more resistant to forgery and removal attacks \cite{zhao2025invisible}.

\vspace{-10pt}
\paragraph{Forgery and Removal Attacks against Watermarks.}

Forging and removing watermarking has been an important research area to expose vulnerabilities in watermarking schemes, and therefore lead to their improvement. Zhao et al. \cite{zhao2025invisible} had shown that many watermarks can be removed by simply noising and then denoising a watermarked image using a diffusion model; however, their approach was not successful against the Tree-Ring \cite{wen2023tree} watermarking method. Yang and Ci et al. \cite{yang2024steganalysis} showed that most watermarking schemes leave distinct textural patterns in the image, which can be found by averaging multiple watermarked images. A concurrent work, \cite{muller2024black} uses an auxiliary diffusion model to ensure the inverted initial noises from a watermarked and non-watermarked image are closely aligned. 
WAVES \cite{anwaves} benchmarked different watermarking and attack methods to judge their effectiveness. They also proposed a set of attacks to evaluate the robustness of various watermarking schemes. Saberi et al. \cite{saberi2023robustness} proposed training a proxy watermarked image classifier to classify whether an image is watermarked or not. They conducted a progressive gradient-descent-based adversarial attack on the model and showed that the perturbations were transferrable to a black-box detection model. 
Liu et al. \cite{liu2024image} proposed regenerating an image similar to watermarked images from clean Gaussian noise so as to remove the watermark. Lukas et al. \cite{lukas2023leveraging} had proposed using differentiable surrogate keys to learn attack parameters such that they can remove traces of watermarked keys from the image. This assumes access to not only a surrogate key generator but also a copy of the generative model. 

Unlike \cite{saberi2023robustness,yang2024steganalysis}, we do not require access to multiple watermarked images. They assume access to images not only from the same watermarking method but also from the same secret key, making these attacks less practical against some systems \citep{arabi2024hidden}. We can run our attack using just one watermarked image. Furthermore, unlike \cite{muller2024black,lukas2023leveraging}, we do not assume any access to a denoising diffusion model or a proxy version of it. Lastly, unlike \cite{zhao2025invisible}, which was not successful in removing the Tree-Ring watermark, we show that our method can do it.


\section{Preliminaries}

Diffusion models \citep{song2020denoising} such as Stable Diffusion (SD) \citep{rombach2022high_LDM} and Imagen \citep{saharia2022photorealistic} learn a mapping from an initial random noise state $\mathbf{z}_T \sim \mathcal{N}(0, \mathbf{I})$ to a clean image space $\mathbf{z}_{0} \sim p_{\text{data}}$. This is done by iteratively applying a learned denoising network $\bm{\epsilon}_{\theta}$ such as U-Net or DiT. Popularly used models \cite{rombach2022high_LDM} compress the image space to a lower-dimensional representation space using a variational autoencoder (an encoder $\mathbf{\mathcal{E}_{\theta}}$ and decoder $\mathbf{\mathcal{D}_{\theta}}$) to reduce the computations required for generating an image.

Using the learned noise estimator network $\bm{\epsilon}_{\theta}$, DDIM's sampling process~\cite{song2020denoising} computes the previous state $\mathbf{z}_{t-1}$ from $\mathbf{z}_{t}$ as follows: 
\begin{equation}
    \mathbf{z}_{t-1} =\sqrt{\frac{\Bar{\alpha}_{t-1}}{\Bar{\alpha}_t}}\mathbf{z}_t - (\sqrt{\frac{1}{\Bar{\alpha}_{t-1}}-1 } - \sqrt{\frac{1}{\Bar{\alpha}_{t}}-1 })\bm{\epsilon}_{\theta}(\mathbf{z}_t, t, \bm{e}_\text{p}),
    \label{equation:x_t-1}
\end{equation}
where $\beta_t$ is defined by the noise scheduler and $\Bar{\alpha}_t = \prod_{i=1}^{t}(1-\beta_{i})$. 


DDIM inversion \cite{mokady2023null,dhariwal2021diffusion,song2020denoising} is a process to invert a clean sample $\mathbf{z}_0$ to reconstruct its initial noise state $\mathbf{z}_T$ based on the assumption that $\mathbf{z}_{t-1} - \mathbf{z}_{t} \approx \mathbf{z}_{t+1} - \mathbf{z}_{t}$. This allows us to estimate $\mathbf{z}_{t+1}$ from $\mathbf{z}_t$ using the formula, 
\begin{equation}
    \mathbf{z}_{t+1} = \sqrt{\Bar{\alpha}_{t+1}}\mathbf{z}_0 + \sqrt{1 - \Bar{\alpha}_{t+1}}\bm{\epsilon}_{\theta}(\mathbf{z}_t, t).
\end{equation}


\begin{figure*}
    \centering
    \includegraphics[width=0.85\linewidth]{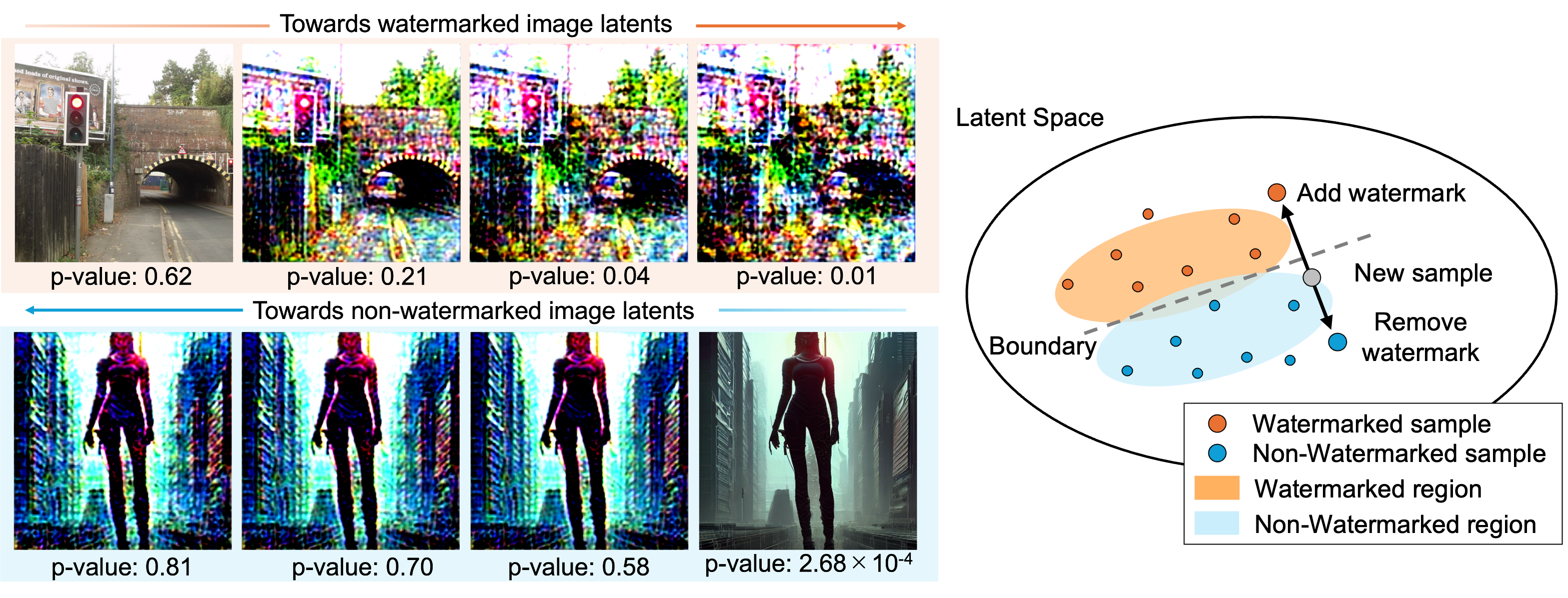}
    \caption{Motivation for the attack. There exists a latent direction/region pertaining to watermarked latents from a specific secret key in the clean image latent space. The further we traverse in the relevant direction, the stronger the attack becomes. Our method exploits this vulnerability while better preserving the image content. See Appendix Figure \ref{fig:directions_appendix} for more examples. }
    \label{fig:latent_result}
\end{figure*}

\subsection{Watermarking Scheme in Diffusion Models}

Most watermarking schemes in diffusion models consist of embedding a secret key $k \in \mathcal{K}$ in the initial noise $\mathbf{z}_T$ used to generate an image. This is done in a manner such that the initial noise does not deviate significantly from a standard Gaussian distribution $\mathcal{N}(0, \mathbf{I})$. The standard diffusion denoising process is followed to convert the new initial noise to a clean watermarked image $\mathbf{x}^{(w)}$ that will be free of visible perturbations. 

During detection, DDIM inversion is used to estimate the initial noise sample $\mathbf{z}_{T}'$ from the clean sample $\mathbf{x}_0$. Once the initial noise is recovered, the key pattern (if any) is extracted and matched with the set of secret keys the model owner used. It is important to note here that the inversion process is done using an empty prompt since the model owner does not keep track of the images that were generated using its model and thus the prompts that might have been used.  

This process allows a model owner to watermark an image without the need for making any modification to the diffusion model architecture or weights. Additionally, this has been shown to be effective against image transformations, which can affect the watermark signal \cite{wen2023tree,ci2024ringid,arabi2024hidden}.

\section{Watermarking Attack Techniques}

In this section, we introduce our watermarking attack techniques.
We start by defining the threat model that we are considering in Section \ref{sec:threat_model}, followed by explaining the motivation for our approach on how to forge or remove watermarks into/from given images in Section \ref{sec:motivation}, and finally we describe the adversarial attack itself in Section \ref{sec:attack_forge}.

\subsection{Threat Model}
\label{sec:threat_model}

We consider two parties, the model owner and the attacker. The model owner owns a generative model and controls the generation such that it outputs watermarked images. 
The attacker is a party with malicious intentions that wants to tamper with the watermarking system. One type of attacker may wish to \textit{forge} the watermark into an unrelated image, to falsely claim that a harmful image was generated by the said model owner. 
Another type of attacker may attempt to \textit{remove} the watermark pattern from a previously watermarked image. This could be done to falsely claim ownership or to spread misinformation by concealing the image’s synthetic origin, making it harder for it to be detected as AI-generated content (e.g., to create deepfakes).
Formally: 
\begin{itemize}
    \item Model Owner (Generation Phase): The model owner owns a diffusion model $\bm{\epsilon}_{\theta}$ and uses a random secret key $k$ from a set of secret keys $\mathcal{K}$ to generate a watermarked image $\mathbf{x}^{(w)}$.
    \item Attacker: The attacker wishes to use a single watermarked image $\mathbf{x}^{(w)}$ to falsely watermark a clean image $\mathbf{x}^{(c)}$. Alternatively, they may wish to remove the watermark in $\mathbf{x}^{(w)}$. The attacker does not have access to the model $\bm{\epsilon}_{\theta}$ used by the model owner or the secret key $k$ that was embedded by the model owner. The attacker can get access to a watermarked sample by sampling the model, but sampling more times is likely to lead to images with different watermarks. 
    \item Model Owner (Detection Phase): The model owner is asked to verify whether or not an image provided by the attacker was generated using their diffusion model $\bm{\epsilon}_{\theta}$ by matching the extracted key to their set of secret keys $\mathcal{K}$. 
\end{itemize}


\subsection{Motivation}
\label{sec:motivation}

Our approach is based on the intuition that the mapping from generated images to initial noise is inherently many-to-one, as the same initial noise sample can produce many different images when denoised using different prompts. In the case of watermark detection, the DDIM inversion process is performed using an empty prompt, meaning that all images generated with a specific secret key, regardless of the original text prompt, are expected to be inverted to recover that key. Based on this, we hypothesize that within the clean sample latent space, there exists a region that consistently maps to an initial noise pattern corresponding to the key.
Our method focuses on exploiting this vulnerability: i.e., if we can successfully embed our non-watermarked sample in the watermarked region, we will be able to falsely claim that a non-watermarked image is watermarked. And vice versa, if we are able to push a watermarked image away from this region, we will be able to falsely claim that it is not watermarked.

\begin{figure}
    \centering
    \includegraphics[trim={2cm 2cm 2cm 2cm},clip,width=\textwidth,width=\linewidth]{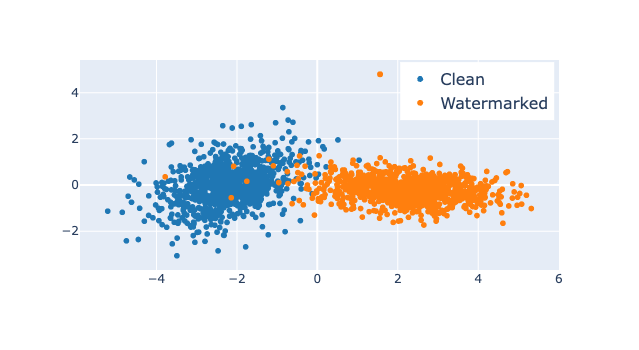}
    \caption{Two-dimensional visualization of the latent space showing the linear separability of watermarked and non-watermarked images. The horizontal axis is obtained by linear discriminant analysis (LDA), while the vertical axis is a random projection.}
    \label{fig:latents_tsne}
\end{figure}



To show that such a region exists, we start by designing a simple experiment. If there exists a latent region pertaining to a different set of watermarked images from each secret key, then we should be able to find latent directions that can lead randomly sampled latent vectors into these regions and thus be authenticated as being watermarked. To demonstrate this, we sampled 1000 watermarked images from a specific watermark and the same number of non-watermarked images. We trained a linear support vector machine (SVM) model on this dataset to find such a direction \cite{colbois2021use}. We observed that simply traversing the latent space in the direction normal to the learned hyperplane could lead a non-watermarked sample towards being classified as watermarked, as shown in Figure~\ref{fig:latent_result}. And vice versa, we can remove the watermark from a watermarked image by traversing the latent space in the opposite direction. We formally define this region in Definition \ref{def:watermark_region}. The linear SVM model yields an accuracy of 100\% on the sampled images, and we also demonstrate the linear separability of the two classes in Figure \ref{fig:latents_tsne}. While this attack in itself is not close to a real-world scenario as we used multiple watermarked samples from the same key and we could not control the semantic content and quality of the corresponding image, this preliminary experiment motivates our novel attack strategy.

\begin{definition}[Watermark Region]\label{def:watermark_region}
For a watermarking method $\mathcal{W}$, we define a watermark region as a region in the clean latent space of the latent diffusion model, as, 
\begin{align*}
Z_0^{(w)}(\mathcal{W},k) &= \Bigl\{ \mathbf{z}_0 \in Z_0 \mid \mathcal{M}_{\mathcal{W}}(\mathcal{I}^{-}(\mathbf{z}_0), k) < \tau \Bigr\},
\end{align*}
where $Z_0$ represents the clean latent space at $t=0$, $\mathcal{I}^{-}$ is the DDIM inversion process, $\mathcal{M}_{\mathcal{W}}$ is the matching function used to verify the presence of a particular key, $\tau$ is the operating threshold of the watermarking scheme, and $k$ is the secret key that was embedded. 

This represents all the points in the clean latent space that lead to the key $k$ that was embedded in the initial noise latent vector when inverted. 
\end{definition}

\subsection{Imperceptible Attack Against Watermarks}
\label{sec:attack_forge}

\paragraph{Forgery Attack.}

Based on the above intuition, to forge the watermark, the attacker needs to adversarially perturb a non-watermarked image so that it gets embedded in the watermarked region of the clean latent space. However, finding and defining this region is not straightforward and requires multiple generations from the same initial noise vector that has the secret key embedded in it. This is not feasible in scenarios where a model owner can regenerate the key every time~\cite{arabi2024hidden}. We instead show that it is sufficient to utilize only one watermarked image for guidance such that if we minimize the distance between the latent representations of the non-watermarked image and a watermarked image, we will be able to successfully guide the non-watermarked image to be embedded in the watermarked region and thus forge the watermark. We can utilize the encoder of an off-the-shelf VAE $\mathcal{E}_\phi$ that was trained on a similar dataset (which can be different from the VAE of the diffusion model $\mathcal{E}_\theta$), to adversarially perturb a non-watermarked image.



The objective for finding the perturbation is defined as:
\begin{equation}
    \min_{\bm{\delta}}  \| \mathcal{E}_{\phi}(\mathbf{x}^{(c)}+ \bm{\delta}) - \mathcal{E}_{\phi}(\mathbf{x}^{(w)}) \|_2 + \lambda \|\bm{\delta}\|_2,
    \label{eq:loss_forgery}
\end{equation}
where $\bm{\delta}$ is the adversarial perturbation and $\lambda$ controls the trade-off between the success of the attack and image content preservation. We saw that this objective works better in practice as compared to finding perturbations using progressive gradient descent (PGD) \cite{kurakin2016adversarial,kurakin2018adversarial}. Additionally, we can devise a loss function that only perturbs the low-frequency regions of the image by optimizing in the frequency domain instead of the RGB domain. We present experimental results to show these settings in the Appendix.

\begin{figure*}[tbh]
    \centering
    \begin{subfigure}[t]{0.2\linewidth}
        \centering
        \includegraphics[width=\linewidth]{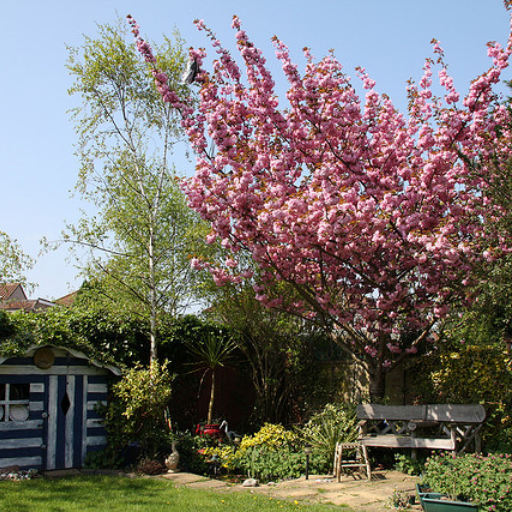}{}
    \end{subfigure}
    ~
    \begin{subfigure}[t]{0.2\linewidth}
        \centering
        \includegraphics[width=\linewidth]{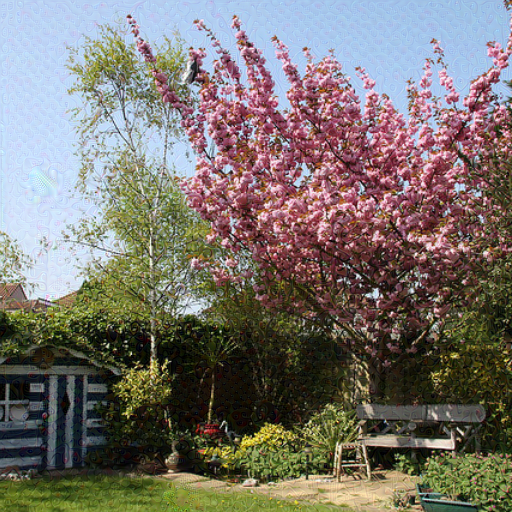}{}
    \end{subfigure}
    ~
    \begin{subfigure}[t]{0.2\linewidth}
        \centering
        \includegraphics[width=\linewidth]{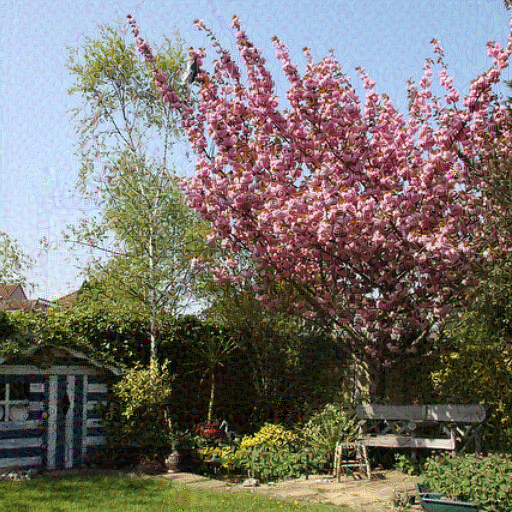}{}
    \end{subfigure}
    ~
    \begin{subfigure}[t]{0.2\linewidth}
        \centering
        \includegraphics[width=\linewidth]{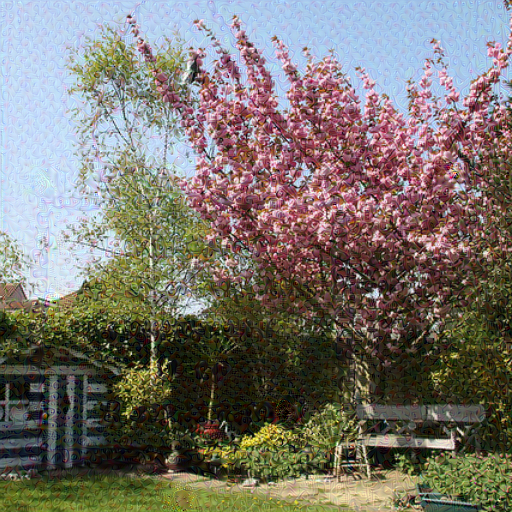}{}
    \end{subfigure}
    \\
    \begin{subfigure}[t]{0.2\linewidth}
        \centering
        \includegraphics[width=\linewidth]{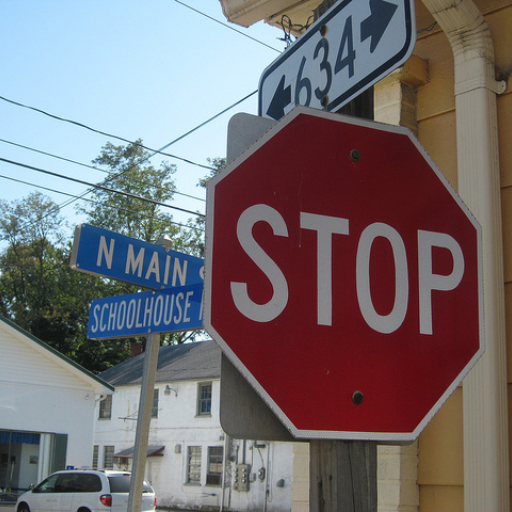}\\
        Non-watermarked
    \end{subfigure}
    ~
    \begin{subfigure}[t]{0.2\linewidth}
        \centering
        \includegraphics[width=\linewidth]{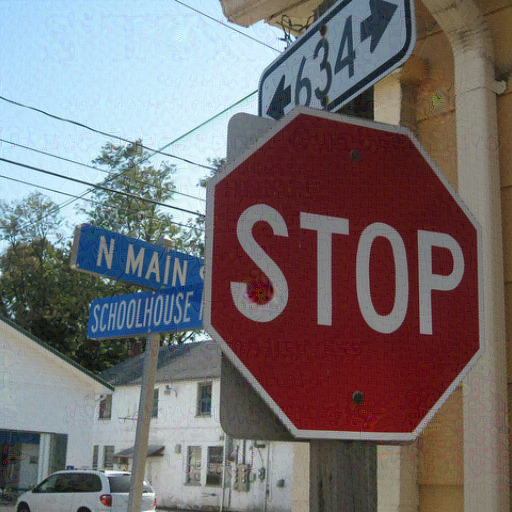}\\
        {$\lambda=5\times 10^4$}
    \end{subfigure}
    ~
    \begin{subfigure}[t]{0.2\linewidth}
        \centering
        \includegraphics[width=\linewidth]{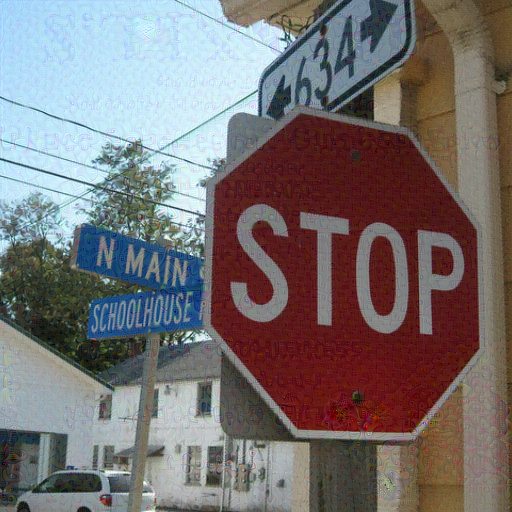}\\
        {$\lambda=2\times 10^4$}
    \end{subfigure}
    ~
    \begin{subfigure}[t]{0.2\linewidth}
        \centering
        \includegraphics[width=\linewidth]{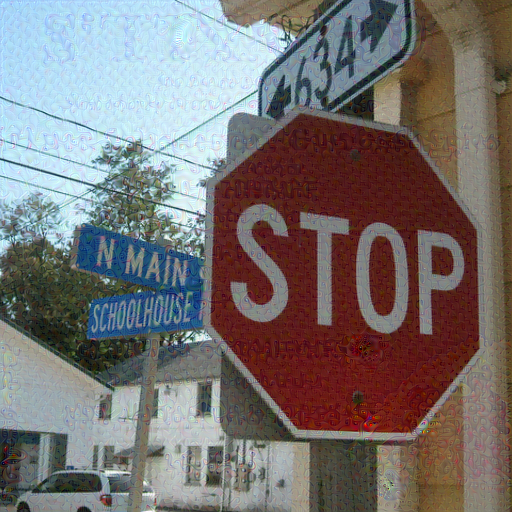}\\
        {$\lambda=1\times 10^4$}
    \end{subfigure}
    
    \caption{Examples showing successful watermark forgery attacks on the Tree-Ring watermarking method with different hyperparameter $\lambda$ values. The hyperparameter $\lambda$ controls the trade-off between ASR and the amount of perturbations we introduce.  }
    \label{fig:tree-rings-examples}
\end{figure*}

\paragraph{Removal Attack.}

We can use a similar approach to remove the watermark by adversarially perturbing a watermarked image such that we can ensure that it gets outside of the vulnerable watermarked region. In this case, we propose using a plain image with all values equal to the mean of the watermarked image $\mathbf{x}^{(w)}$ instead of using a camera captured non-watermarked image as (1) this is naturally non-watermarked and does not require a real image for guidance and (2) real images contain their own high frequency information, which can lead to higher perturbations in the optimized adversarial image. 
We can summarize the removal objective as, 
\begin{equation}
    \min_{\bm{\delta}}  \| \mathcal{E}_{\phi}(\mathbf{x}^{(w)}+ \bm{\delta}) - \mathcal{E}_{\phi}(\bm{\mu}_{\mathbf{x}^{(w)}}) \|_2 + \lambda \|\bm{\delta}\|_2,
    \label{eq:loss_removal}
\end{equation}
where $\bm{\mu}_{\mathbf{x}^{(w)}}$ is the plain image with all values equal to the mean of the watermarked image $\mathbf{x}^{(w)}$. In Appendix \ref{app:remove_real}, we show experimental results to justify this design choice by showing that using real images leads to worse performance.




\section{Experimental Results} 
\label{sec:results}

In this section, we showcase the effectiveness of our approach. We consider two attack scenarios, where the attacker has access to either (i) the VAE of the watermarked diffusion model or (ii) a proxy VAE that was trained on a similar dataset.

\begin{table*}[]
    \centering
    \caption{Trade-off between attack success rate (ASR) and imperceptibility of the attack as judged using the $l_2$, $l_{\infty}$ distances and the Learned Perceptual Image Patch Similarity (LPIPS) \cite{zhang2018perceptual}, Structural Similarity Index (SSIM) \cite{wang2004image}, Peak Signal to Noise Ratio (PSNR) and Fréchet Inception Distance (FID) \cite{heusel2017gans} metrics, when falsely watermarking non-watermarked images from the COCO2017 dataset using the VAE from SDv1.4 for optimization. Hyperparameter $\lambda$ controls the trade-off between ASR and the amount of perturbation we introduce. }
    \resizebox{0.71\textwidth}{!}{
\begin{tabular}{ccc|ccccccc}
\hline 
Method & Model & $\lambda$ & ASR & $l_2$ & $l_{\infty}$ & LPIPS & SSIM  & PSNR & FID \\ \hline 
\multirow{6}{*}{ Tree-Ring \cite{wen2023tree} } & \multirow{3}{*}{ SDv1.4 } & $5\times 10^4$ & 78.65 & 33.90 & 0.69 & 0.17 & 0.89 & 34.32 & 41.27 \\ 
& & $2\times 10^4$ & 86.93 & 48.42 & 0.89 & 0.26 & 0.82 & 31.20 & 61.18 \\
& & $1\times 10^4$ & 91.06 & 63.22 & 1.10 & 0.33 & 0.76 & 28.87 & 81.27 \\
\cline{2-10}
\multirow{3}{*}{ } & \multirow{3}{*}{ SDv2.0 } & $5\times 10^4$ & 79.89 & 34.09 & 0.69 & 0.17 & 0.88 & 34.26 & 41.22 \\
& & $2\times 10^4$ & 90.72 & 48.83 & 0.91 & 0.26 & 0.82 & 31.11 & 61.75 \\
& & $1\times 10^4$ & 93.81 & 63.78 & 1.08 & 0.34 & 0.76 & 28.78 & 80.54 \\
\hline 
\multirow{6}{*}{ RingID \cite{ci2024ringid} } & \multirow{3}{*}{ SDv1.4 } & $5\times 10^4$ &  100.0 & 38.45 & 0.68 & 0.20 & 0.87 & 33.21 & 49.97 \\
& & $2\times 10^4$ & 100.0 & 55.20 & 0.86 & 0.30 & 0.80 & 30.06 & 75.55 \\
& & $1\times 10^4$ & 100.0 & 73.08 & 1.03 & 0.38 & 0.73 & 27.63 & 101.45 \\ \cline{2-10}
\multirow{3}{*}{ } & \multirow{3}{*}{ SDv2.0 } & $5\times 10^4$ & 100.0 & 37.31 & 0.66 & 0.19 & 0.87 & 33.48 & 47.87 \\
& & $2\times 10^4$ & 100.0 & 53.94 & 0.84 & 0.29 & 0.80 & 30.27 & 72.08 \\
& & $1\times 10^4$ & 100.0 & 71.53 & 1.00 & 0.37 & 0.73 & 27.82 & 98.12 \\
\hline 
\multirow{6}{*}{ WIND \cite{arabi2024hidden} } & \multirow{3}{*}{ SDv1.4 } & $5\times 10^4$ & 97.56 & 38.82 & 0.70 & 0.20 & 0.87 & 33.11 & 54.41 \\
& & $2\times 10^4$ & 97.56 & 56.13 & 0.89 & 0.29 & 0.80 & 29.88 & 82.68 \\
& & $1\times 10^4$ & 97.56 & 74.66 & 1.06 & 0.38 & 0.73 & 27.38 & 111.84 \\ \cline{2-10}
&  \multirow{3}{*}{ SDv2.0 } & $5\times 10^4$ & 100.0 & 37.47 & 0.67 & 0.19 & 0.87 & 33.45 & 49.02 \\
& & $2\times 10^4$ & 100.0 & 54.18 & 0.84 & 0.28 & 0.80 & 30.23 & 74.69 \\
& & $1\times 10^4$ & 100.0 & 71.86 & 0.99 & 0.37 & 0.74 & 27.78 & 101.33 \\
\hline 
\multirow{6}{*}{ Gaussian Shading \cite{yang2024gaussian}} & \multirow{3}{*}{ SDv1.4 } & $5\times 10^4$ & 96.85 & 37.27 & 0.70 & 0.19 & 0.87 & 33.48 & 46.93 \\
& & $2\times 10^4$ &  96.96 & 54.00 & 0.88 & 0.29 & 0.80 & 30.21 & 69.50 \\
& & $1\times 10^4$ & 96.96 & 71.97 & 1.05 & 0.37 & 0.73 & 27.64 & 93.85 \\ \cline{2-10}
&  \multirow{3}{*}{ SDv2.0 } & $5\times 10^4$ & 100.0 & 36.78 & 0.66 & 0.19 & 0.87 & 33.60 & 46.20 \\
& & $2\times 10^4$ & 100.0 & 52.99 & 0.85 & 0.29 & 0.80 & 30.42 & 69.35 \\
& & $1\times 10^4$ & 100.0 & 69.83 & 1.02 & 0.37 & 0.74 & 28.02 & 93.04 \\ 
\hline 
\end{tabular}
    }
    \label{tab:false}
\end{table*}

\paragraph{Experimental Setup.}
We consider two diffusion models, namely, Stable Diffusion v1.4 (SDv1.4) and Stable Diffusion v2.0 (SDv2.0), to generate images of size $512\times 512$. 
We generate watermarked images using the prompts available in the \textit{Gustavosta/Stable-Diffusion-Prompts}
. When generating reference watermarked images for forgery attacks, we use simpler prompts from the \textit{runwayml-stable-diffusion-v1-5-eval-random-prompts} 
 dataset as the resultant images contain more visible watermark patterns/signal due to lower amounts of high-frequency information. 
We use the COCO2017 validation dataset \cite{lin2014microsoft} to obtain images without watermarks. We randomly select 200 pairs of watermarked and non-watermarked images for both forgery and removal attacks. We utilize the VAE from SDv1.4 to remove/forge the watermark generated from both SDv1.4 and SDv2.0. We report results on the following publicly available watermarking systems that embed a key in the initial noise space, namely, Tree-Ring \cite{wen2023tree}, RingID \cite{ci2024ringid}, WIND \cite{arabi2024hidden}, and Gaussian Shading \cite{yang2024gaussian}. See Appendix \ref{app:implementation} for further details.  

\paragraph{Evaluation Metrics.} We consider a forgery attack to be successful if the $p$-value statistical test between the extracted key and the secret key embedded in the reference watermarked image yields a $p$-value less than 0.05. For removal, we consider it a success if the $p$-value is greater than 0.05 with respect to the key in the original watermarked image. For Gaussian Shading, which uses a bit sequence, we compute the binary bit accuracy between the embedded and recovered keys. The threshold for this is computed at a false positive rate of $10^{-6}$ as set by the original authors. We report the average success rate across 200 examples while randomly drawing new non-watermarked and watermarked samples each time. Even though we test our attack in a black-box scenario, we do not make multiple attempts and consider it unsuccessful if the first attempt fails. We additionally report the $l_2$, $l_{\infty}$ distances and the Learned Perceptual Image Patch Similarity (LPIPS) \cite{zhang2018perceptual}, Structural Similarity Index (SSIM) \cite{wang2004image}, Peak Signal to Noise Ratio (PSNR) and Fréchet Inception Distance (FID) \cite{heusel2017gans} metrics between the original image and the adversarially perturbed image to assess the extent to which we alter the original image. 

\paragraph{Results - Forgery.} We summarize the results in Table \ref{tab:false} and show visual examples in Figure \ref{fig:tree-rings-examples}. Our method achieves almost perfect scores against the RingID, WIND, and Gaussian Shading watermarks even in the black-box setting where we utilize a proxy VAE. We also report high ASR against the Tree-Ring watermarking scheme, $\approx 85\%$ at higher perturbation budgets. We showcase more qualitative examples in the Appendix \ref{app:visual}. 

\begin{figure*}[tbh]
    \centering
    \begin{subfigure}[t]{0.2\linewidth}
        \centering
        \includegraphics[width=\linewidth]{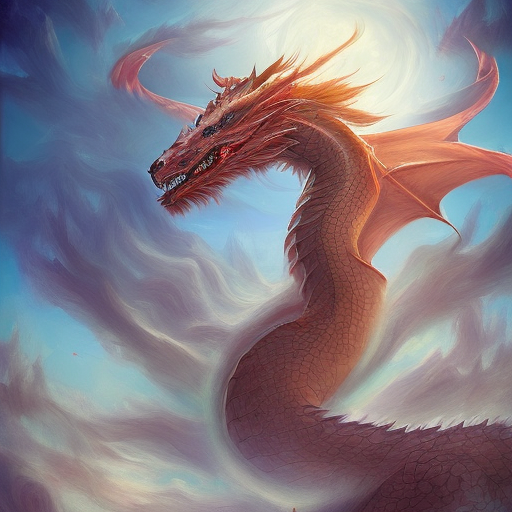}{}
    \end{subfigure}
    ~
    \begin{subfigure}[t]{0.2\linewidth}
        \centering
        \includegraphics[width=\linewidth]{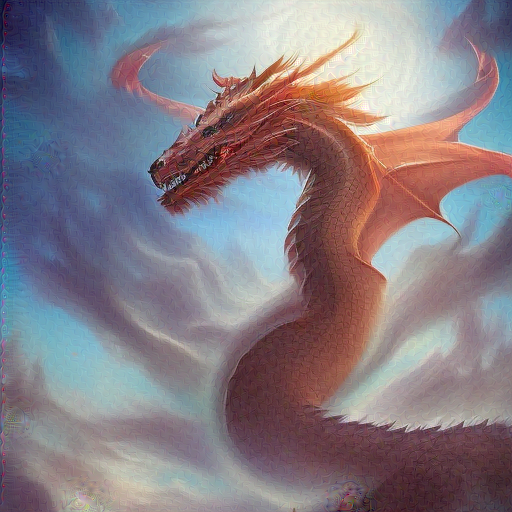}{}
    \end{subfigure}
    ~
    \begin{subfigure}[t]{0.2\linewidth}
        \centering
        \includegraphics[width=\linewidth]{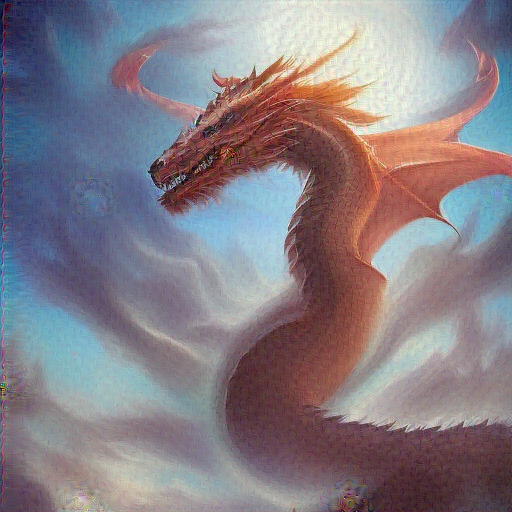}{}
    \end{subfigure}
    ~
    \begin{subfigure}[t]{0.2\linewidth}
        \centering
        \includegraphics[width=\linewidth]{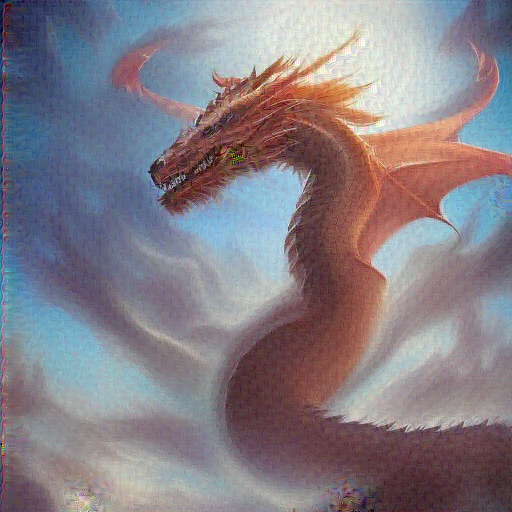}{}
    \end{subfigure}
     \\
    \begin{subfigure}[t]{0.2\linewidth}
        \centering
        \includegraphics[width=\linewidth]{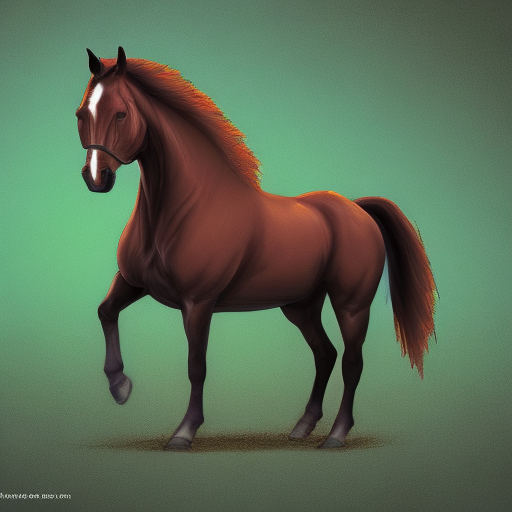}\\
        Original Watermarked
    \end{subfigure}
    ~
    \begin{subfigure}[t]{0.2\linewidth}
        \centering
        \includegraphics[width=\linewidth]{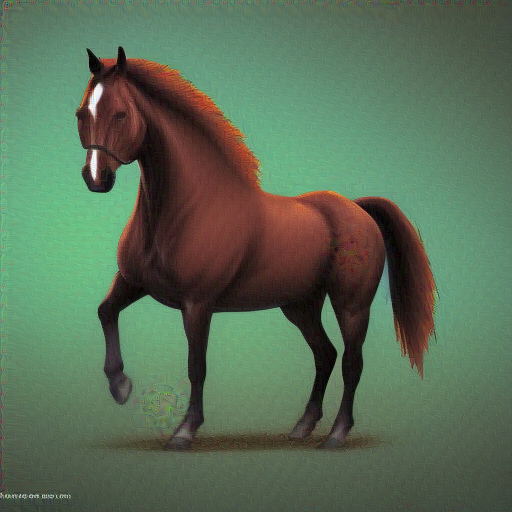}\\
        {$\lambda=5\times 10^4$}
    \end{subfigure}
    ~
    \begin{subfigure}[t]{0.2\linewidth}
        \centering
        \includegraphics[width=\linewidth]{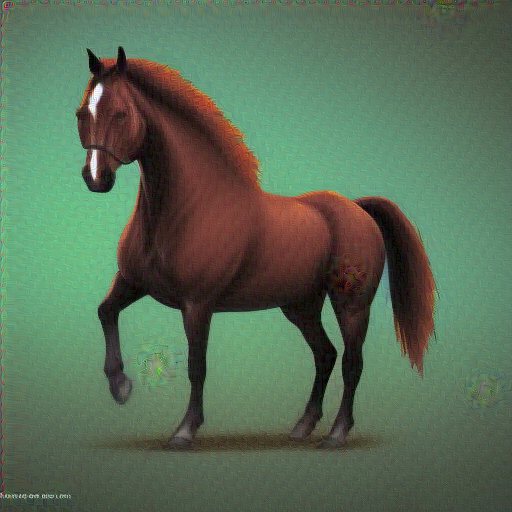}\\
        {$\lambda=2\times 10^4$}
    \end{subfigure}
    ~
    \begin{subfigure}[t]{0.2\linewidth}
        \centering
        \includegraphics[width=\linewidth]{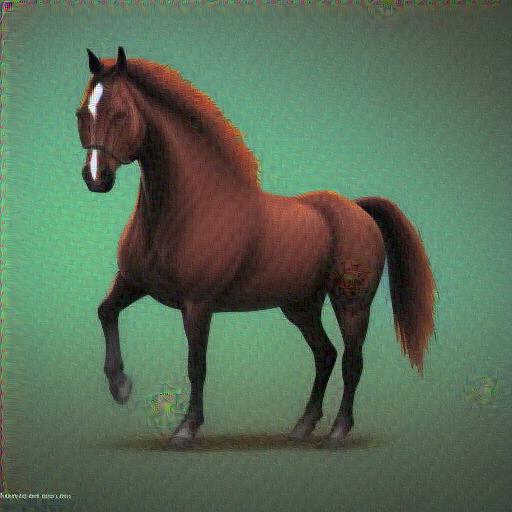}\\
        {$\lambda=1\times 10^4$}
    \end{subfigure}
    
    \caption{Examples showing successful watermark removal attacks on the Tree-Ring watermarking method with different hyperparameter $\lambda$ values. The hyperparameter $\lambda$ controls the trade-off between ASR and the amount of perturbations we introduce. }
    \label{fig:tree-rings-examples_removal}
\end{figure*}

\vspace{-10pt}
\paragraph{Results - Removal.} We report results on removing the watermark in Table \ref{tab:remove} for Tree-Ring and Gaussian Shading and in Appendix Table \ref{tab:remove_wind_ringid} for RingID and WIND watermarking schemes. Although our approach achieves success at removing the Tree-Ring watermark, it was harder to remove the watermark signal from Gaussian Shading, RingID and WIND. As these methods embed a watermark into the entire initial latent noise space, which introduces a larger amount of watermark signal into the image (see Sec. \ref{sec:discussion}). 
We show qualitative results when removing the Tree-Rings watermark in Figure \ref{fig:tree-rings-examples_removal}. 






\begin{table*}[]
\centering
\caption{Trade-off between attack success rate (ASR) of removing the watermark and imperceptibility of the attack as judged using the $l_2$, $l_{\infty}$ distances and the Learned Perceptual Image Patch Similarity (LPIPS) \cite{zhang2018perceptual}, Structural Similarity Index (SSIM) \cite{wang2004image}, Peak Signal to Noise Ratio (PSNR) and Fréchet Inception Distance (FID) \cite{heusel2017gans} metrics. We use the VAE from SDv1.4 for optimization. Hyperparameter $\lambda$ controls the trade-off between ASR and the amount of perturbation we introduce. }
    \resizebox{0.71\textwidth}{!}{
\begin{tabular}{ccc|ccccccc}
\hline 
Method & Model & $\lambda$ & ASR & $l_2$ & $l_{\infty}$ & LPIPS & SSIM  & PSNR & FID \\ \hline 
\multirow{6}{*}{ Tree-Ring \cite{wen2023tree} } & \multirow{3}{*}{ SDv1.4 } & $5\times 10^4$ & 94.21 & 74.82 & 0.94 & 0.19 & 0.84 & 20.72 & 54.16  \\ 
& & $2\times 10^4$ & 97.68 & 87.20 & 1.14 & 0.27 & 0.79 & 20.61 & 75.74  \\
& & $1\times 10^4$ & 98.84 & 100.52 & 1.35 & 0.34 & 0.74 & 20.44 & 94.62   \\
\cline{2-10}
\multirow{3}{*}{ } & \multirow{3}{*}{ SDv2.0 } & $5\times 10^4$ & 95.08 & 88.84 & 1.06 & 0.21 & 0.82 & 19.81 & 59.46  \\
& & $2\times 10^4$ & 97.80 & 100.24 & 1.22 & 0.28 & 0.77 & 19.71 & 81.83  \\
& & $1\times 10^4$ & 98.36 & 112.24 & 1.41 & 0.35 & 0.72 & 19.60 & 97.41 \\
\hline 
\multirow{6}{*}{ Gaussian Shading \cite{yang2024gaussian}} & \multirow{3}{*}{ SDv1.4 } & $5\times 10^4$ & 11.41 & 34.68 & 0.86 & 0.13 & 0.89 & 33.95 & 44.57  \\
& & $2\times 10^4$ & 39.79 & 63.76 & 1.15 & 0.23 & 0.81 & 23.99 & 66.48 \\
& & $1\times 10^4$ & 70.10 & 74.10 & 1.31 & 0.29 & 0.77 & 24.12 & 82.75  \\ 
\cline{2-10}
&  \multirow{3}{*}{ SDv2.0 } & $5\times 10^4$ & 12.23 & 38.21 & 0.98 & 0.13 & 0.88 & 33.17 & 43.24 \\
& & $2\times 10^4$ & 34.73 & 53.55 & 1.19 & 0.21 & 0.83 & 30.26 & 63.89 \\
& & $1\times 10^4$ & 59.13 & 68.73 & 1.39 & 0.27 & 0.77 & 28.13 & 82.59  \\ 
\hline 
\end{tabular}
}
\label{tab:remove}
\end{table*}

\begin{figure}
    \centering
    \begin{subfigure}[t]{0.22\linewidth}
        \centering
        \includegraphics[width=\linewidth]{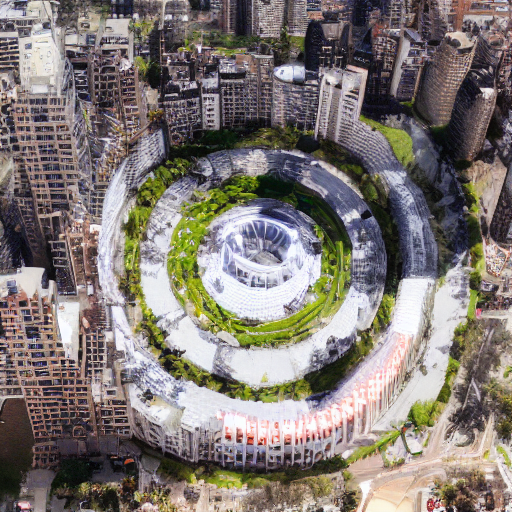}{}
    \end{subfigure}
    \begin{subfigure}[t]{0.22\linewidth}
        \centering
        \includegraphics[width=\linewidth]{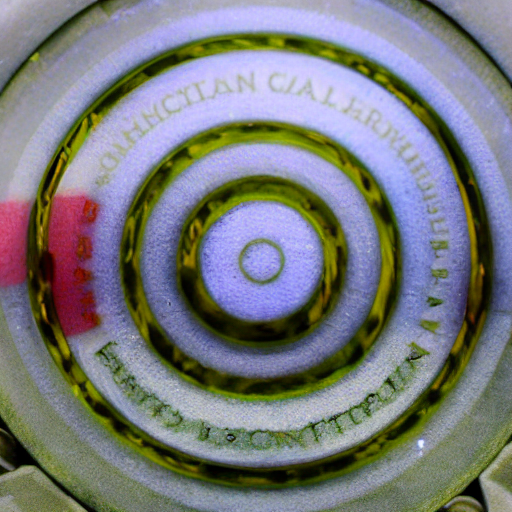}{}
    \end{subfigure}
    \begin{subfigure}[t]{0.22\linewidth}
        \centering
        \includegraphics[width=\linewidth]{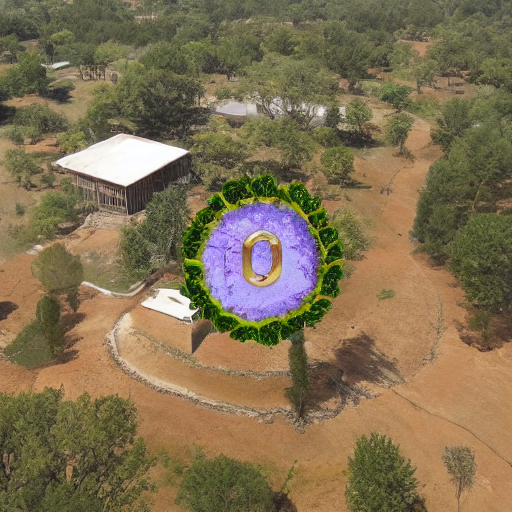}{}
    \end{subfigure}
    \begin{subfigure}[t]{0.22\linewidth}
        \centering
        \includegraphics[width=\linewidth]{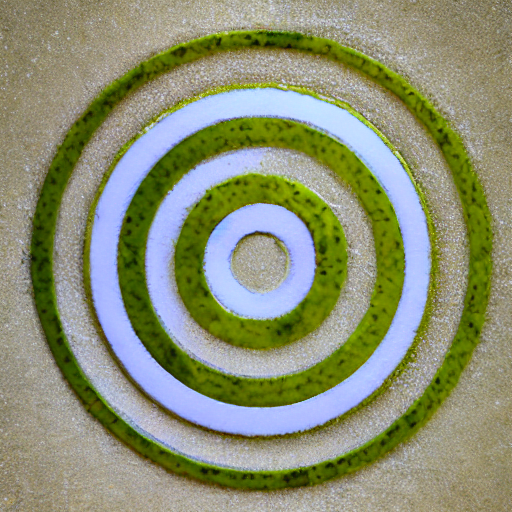}{}
    \end{subfigure}
    \caption{Examples of images generated using RingID, where the watermark signal is visible even in the final generated image.  }
    \label{fig:ringid_bad_examples}
\end{figure}

\section{Discussion and Limitation}
\label{sec:discussion}

\paragraph{Multi-Pattern vs. Single-Pattern Watermarks.}

We found that forging was generally easier for our method than watermark removal across all approaches, particularly for RingID, Gaussian Shading, and WIND watermarks. We suggest that this is because these methods do not merely encode a single pattern like Tree-Ring but encode additional information, such as the model owner's identity or other metadata.
As the number of possible embedded patterns increases, more encoded information is required to correctly identify the pattern. This, in turn, necessitates a stronger signal-to-noise ratio \cite{shannon1949communication}, where noise refers to patterns in the initial noise that are unrelated to our watermark. When a signal is more associated with the embedded watermark, forging at least part of it becomes easier, while completely removing it becomes more difficult.


In contrast, Tree-Ring embeds a simpler pattern, affecting only a portion of the initial noise, and making it harder to forge under a fixed budget.



\paragraph{Baseline Distortion in RingID-type Methods.} While evaluating image quality, we found that even unperturbed images generated by the RingID and WIND methods exhibit some distortion, as shown in Figure~\ref{fig:ringid_bad_examples}. This issue was also discussed in Appendix~C of the RingID manuscript~\cite{ci2024ringid}. The RingID method (and WIND, by incorporating RingID in one of its variants) mitigated these artifacts by embedding a watermark only in a single channel. However, in our evaluation, the distortion is often still noticeable and is more pronounced in the case of simpler prompts. This further highlights the strength of the watermark signal embedded by these techniques, making it harder to remove. 


\paragraph{Alternative Watermarking Approaches.} Other watermarking approaches that add a pattern during the latent decoding phase are less susceptible to attacks such as ours \cite{ci2024wmadapter,fernandez2023stable}. As these methods fine-tune the decoder, the adversary would require access to a similar decoder to attack the system. There is an inherent trade-off here, i.e., these methods are less resistant to image transformations, which was a major advantage of initial noise-based watermarking schemes. 

Other alternatives towards building a more secure watermarking scheme could be to embed a secret message that encodes some information pertaining to the contents of the watermarked image. This would make it harder for an attacker to forge the watermark as they would need to embed a new message, which can only be done if they have access to both the entire diffusion model and the secret message generation method. It would allow a model owner to quickly verify that the contents of the image and recovered secret message match.  



\section{Conclusion}

In this paper, we expose a vulnerability of the Tree-Ring, RingID, WIND, and Gaussian Shading-based watermarking schemes. We show that when the watermark key is embedded in the initial noise, it leads to a latent watermarked region, forming in the denoised latent space. This makes it easier for an attacker to forge the watermark by perturbing the sample so that it lies within this region. We show that a similar approach can also be used for removal by pushing a watermarked latent away from this region. We hope this work motivates future research on improving watermarking systems in the face of adversaries.

{
    \small
    \bibliographystyle{ieeenat_fullname}
    \bibliography{main}
}

\clearpage
\setcounter{page}{1}
\maketitlesupplementary

\begin{figure}[t]
    \centering
    \includegraphics[width=\linewidth]{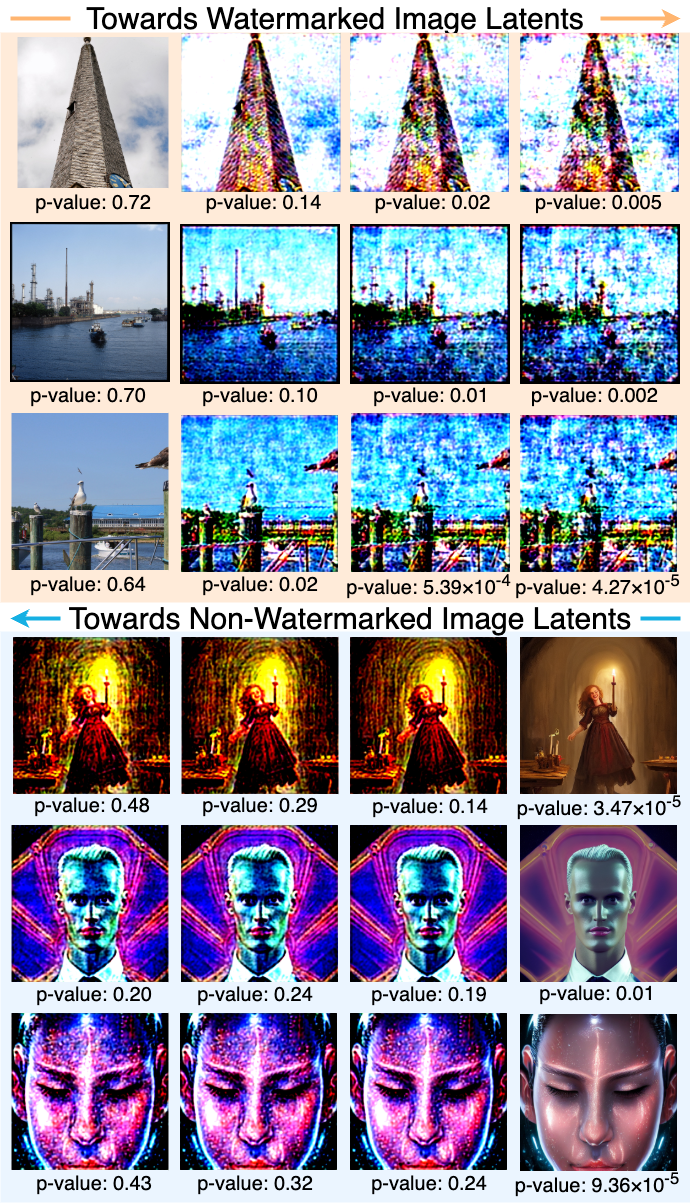}
    \caption{Examples showcasing the idea that there exist latent directions which pertain to watermarking and removal. We learn these directions using a linear SVM and they are the normal to the learned hyperplane. Traversing further along them increases the strength of the attack. }
    \label{fig:directions_appendix}
\end{figure}
\begin{figure*}
    \centering
    \begin{subfigure}[t]{0.15\linewidth}
        \centering
        \includegraphics[width=\linewidth]{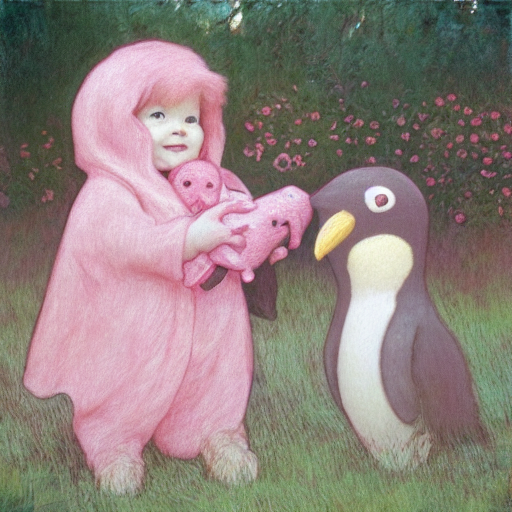}
    \end{subfigure}
    ~
    \begin{subfigure}[t]{0.15\linewidth}
        \centering
        \includegraphics[width=\linewidth]{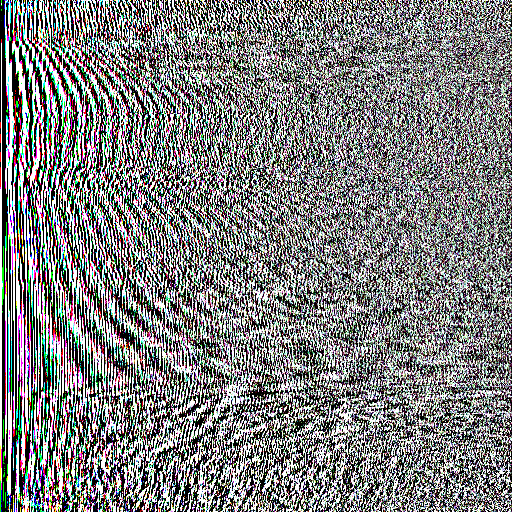}
    \end{subfigure}
    ~
\begin{subfigure}[t]{0.15\linewidth}
        \centering
        \includegraphics[width=\linewidth]{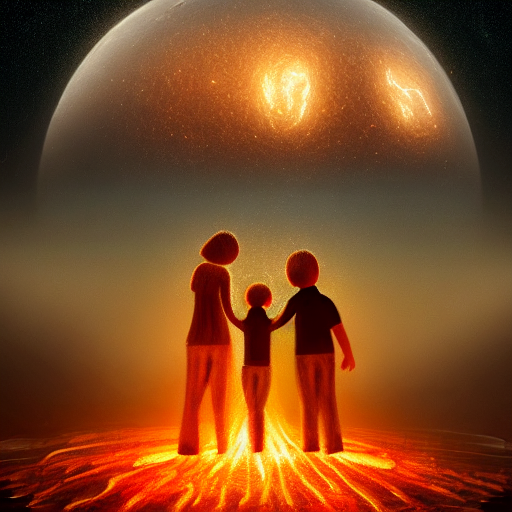}
    \end{subfigure}
    ~
    \begin{subfigure}[t]{0.15\linewidth}
        \centering
        \includegraphics[width=\linewidth]{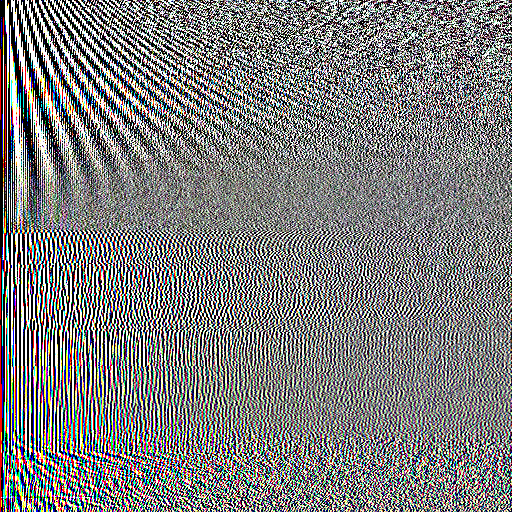}
    \end{subfigure}
    ~
    \begin{subfigure}[t]{0.15\linewidth}
        \centering
        \includegraphics[width=\linewidth]{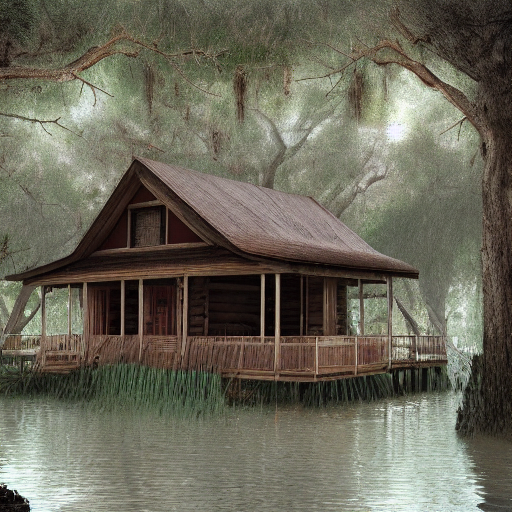}
    \end{subfigure}
    ~
    \begin{subfigure}[t]{0.15\linewidth}
        \centering
        \includegraphics[width=\linewidth]{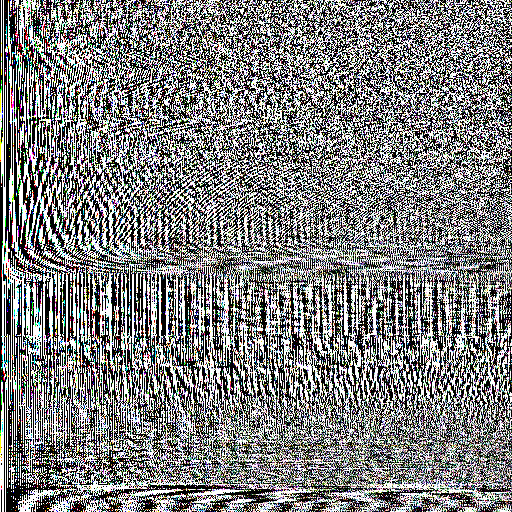}
    \end{subfigure}
    \\
    \begin{subfigure}[t]{0.15\linewidth}
        \centering
        \includegraphics[width=\linewidth]{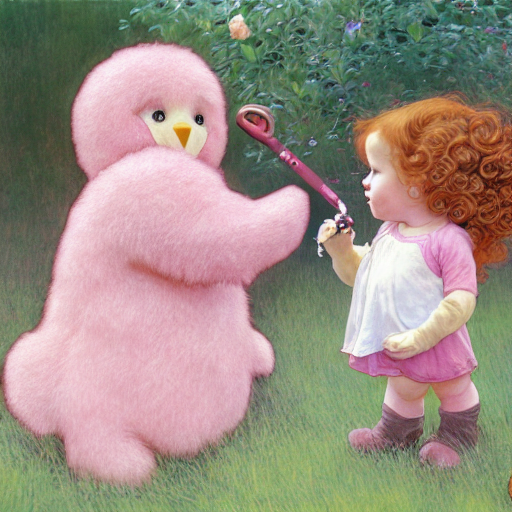}
    \end{subfigure}
    ~
    \begin{subfigure}[t]{0.15\linewidth}
        \centering
        \includegraphics[width=\linewidth]{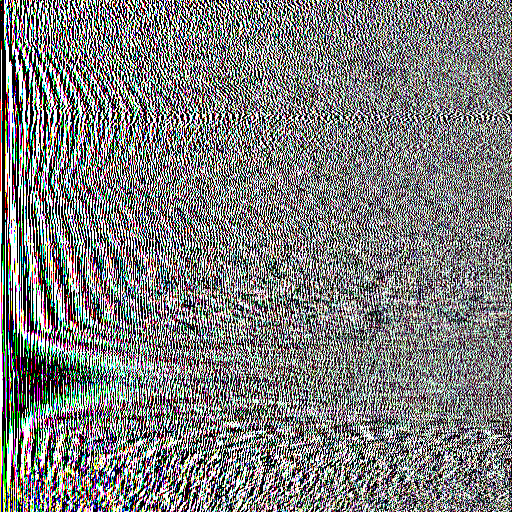}
    \end{subfigure}
    ~
    \begin{subfigure}[t]{0.15\linewidth}
        \centering
        \includegraphics[width=\linewidth]{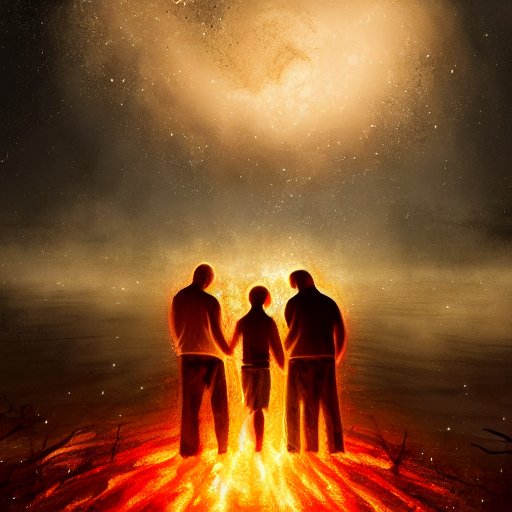}
    \end{subfigure}
    ~
    \begin{subfigure}[t]{0.15\linewidth}
        \centering
        \includegraphics[width=\linewidth]{images/no_w_dct_31.png}
    \end{subfigure}
    ~
    \begin{subfigure}[t]{0.15\linewidth}
        \centering
        \includegraphics[width=\linewidth]{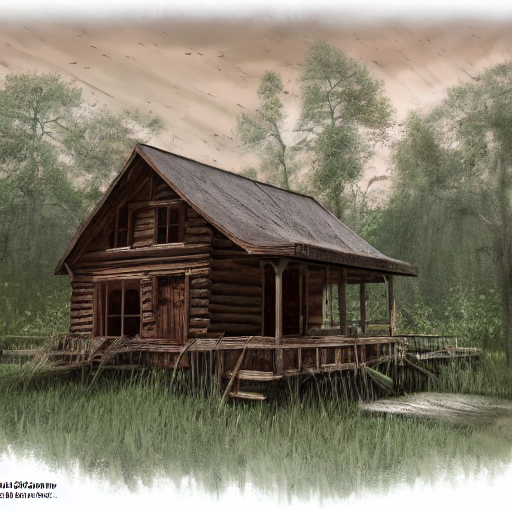}
    \end{subfigure}
    ~
    \begin{subfigure}[t]{0.15\linewidth}
        \centering
        \includegraphics[width=\linewidth]{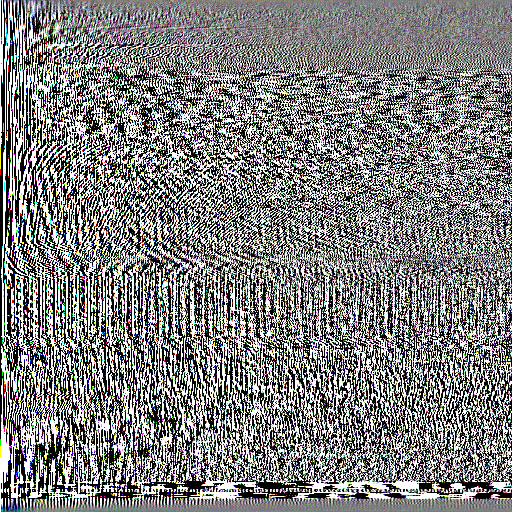}
    \end{subfigure}
    \\
    \begin{subfigure}[t]{0.15\linewidth}
        \centering
        \includegraphics[width=\linewidth]{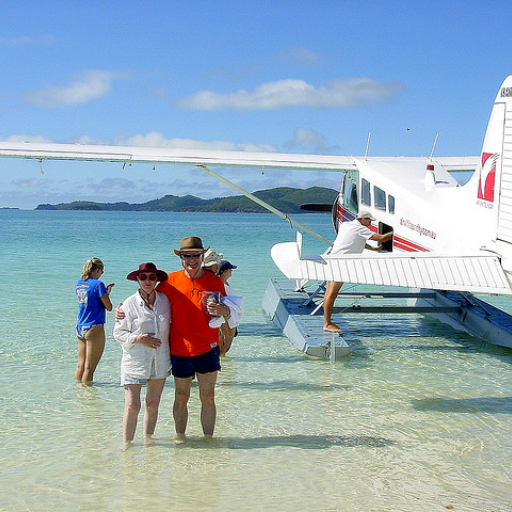}
    \end{subfigure}
    ~
    \begin{subfigure}[t]{0.15\linewidth}
        \centering
        \includegraphics[width=\linewidth]{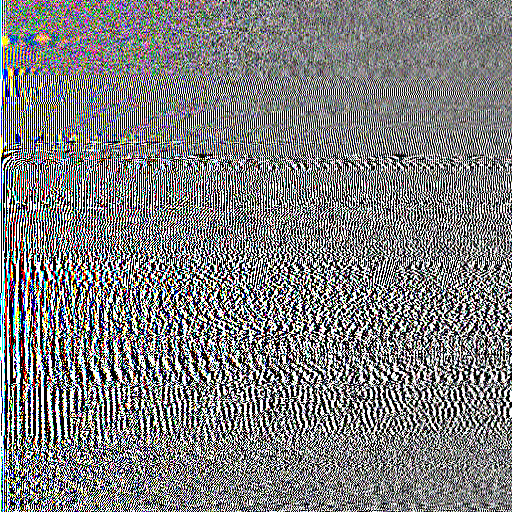}
    \end{subfigure}
    ~
    \begin{subfigure}[t]{0.15\linewidth}
        \centering
        \includegraphics[width=\linewidth]{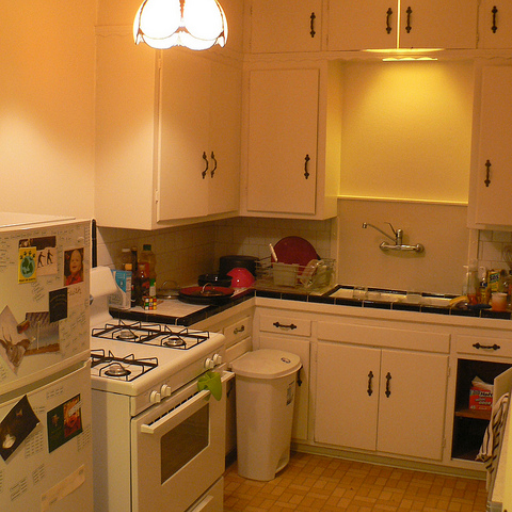}
    \end{subfigure}
    ~
    \begin{subfigure}[t]{0.15\linewidth}
        \centering
        \includegraphics[width=\linewidth]{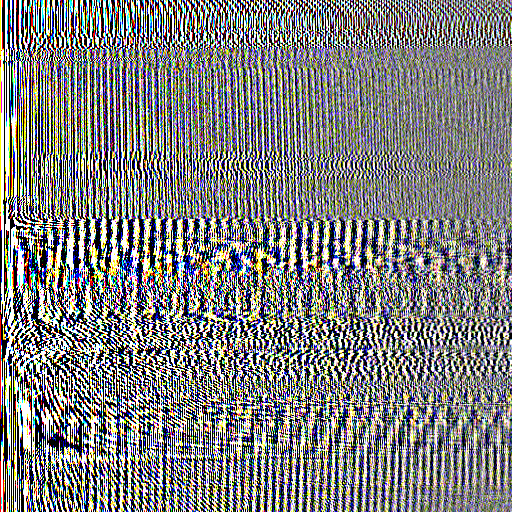}
    \end{subfigure}
    ~
    \begin{subfigure}[t]{0.15\linewidth}
        \centering
        \includegraphics[width=\linewidth]{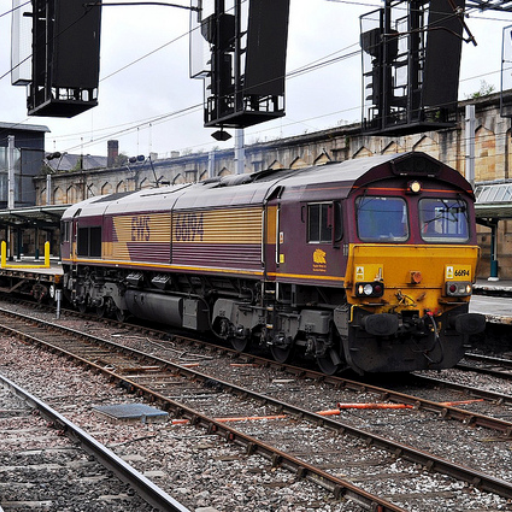}
    \end{subfigure}
    ~
    \begin{subfigure}[t]{0.15\linewidth}
        \centering
        \includegraphics[width=\linewidth]{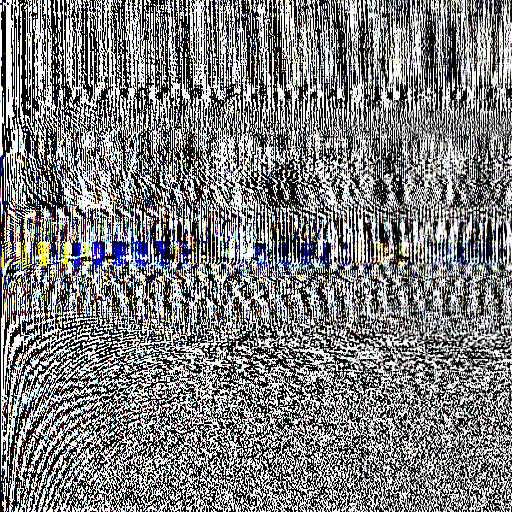}
    \end{subfigure}
    
    \caption{DCT domain representations of watermarked images from the same key (row 1) and non-watermarked images (row 2) generated by SDv1.4 using the same prompts; non-watermarked images from the COCO dataset (row 3).}
    \label{fig:dct}
\end{figure*}

We present the following contents in the Appendix:
\begin{itemize}
    \item More experimental results on watermark removal at different perturbation budgets are presented in Table \ref{tab:remove_wind_ringid}. 
    \item Experimental results on alternative optimization objectives in Section \ref{sec:alternative_loss}.
    \item Experimental results on watermark removal using images with fixed pixel values in Section \ref{app:remove_127.5}.
    \item Experimental results on watermark removal using camera captured non-watermarked images in Section \ref{app:remove_real}.
    \item Implementation details of our method in Section \ref{app:implementation}. 
    \item More visual examples of successful watermark forgery and removal attacks in Section \ref{app:visual}.
    \item Visual examples showcasing the effectiveness of latent directions in forging and removing watermarks in Figure \ref{fig:directions_appendix}. 
    \item Visual examples showing DCT domain representations of watermarked and non-watermarked images in Figure \ref{fig:dct}. 
\end{itemize}

\section{Experimental Results using Alternative Loss Formulations}
\label{sec:alternative_loss}

In this section, we present an alternative design of the adversarial loss function to remove and forge the watermark signal.

\subsection{Using Progressive Gradient Descent}

In our main experiments (Equation \ref{eq:loss_forgery}), we control the amount of perturbations by adding an additional term to our loss function to minimize the perturbations while attacking the watermarking method. In this section, we present an alternative optimization objective by utilizing progressive gradient descent \cite{kurakin2016adversarial,kurakin2018adversarial}, through which we can set an explicit perturbation budget to control the amount of perturbations we introduce. 

The objective for the forgery attack in this case becomes, 
\begin{equation}
    \min_{\bm{\delta}}  \| \mathcal{E}_{\phi}(\mathbf{x}^{(c)}+ \bm{\delta}) - \mathcal{E}_{\phi}(\mathbf{x}^{(w)}) \|_2 \quad \text{s.t.} \quad \|\bm{\delta}\|_\infty \leq \epsilon,
    \label{eq:pgd}
\end{equation}
where $\epsilon$ is the perturbation budget. 

\subsection{Perturbations in the frequency domain}

\begin{figure}
    \centering
    \includegraphics[width=0.5\linewidth]{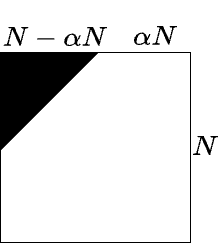}
    \caption{Visualization of the mask we utilized to control the adversarial perturbations such that we add noise only in the higher frequency region of an image. }
    \label{fig:mask}
\end{figure}

Furthermore, we can reduce the perceptible impact by instead of directly perturbing in the RGB space, we convert the image to its frequency domain using the discrete cosine transform (DCT) and perturb only the high-frequency regions in the frequency domain \cite{jia2022exploring,luo2022frequency}. To control the frequency regions we want to alter, we introduce a binary mask $\mathbf{m} \in \{0,1\}^{N\times N}$, which consists of zeros in the upper-left triangle that corresponds to the low-frequency region ending at indices $\lfloor N - \alpha \times N \rfloor$ and ones everywhere else. Here, $\alpha \in [0,1]$ controls the frequency regions we want to perturb (see Figure \ref{fig:mask} for reference on how the mask looks). 

We can write this objective as,
\begin{multline}
    \min_{\bm{\delta}}  \| \mathcal{E}_{\phi}(\text{IDCT}( \text{DCT}(\mathbf{x}^{(c)})+ \mathbf{m}\odot\bm{\delta})) - \mathcal{E}_{\phi}(\mathbf{x}^{(w)}) \|_2 \\ \quad \text{s.t.} \quad \|\bm{\delta}\|_\infty \leq \epsilon.
    \label{eq:freq}
\end{multline}

\begin{figure*}[tbh]
    \centering
    \begin{subfigure}[t]{0.18\linewidth}
        \centering
        \includegraphics[width=\linewidth]{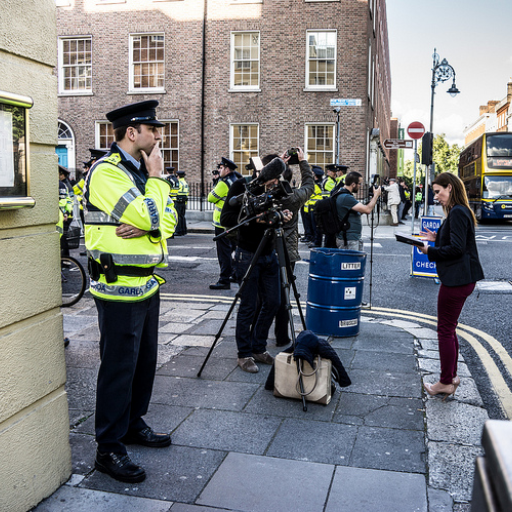}{}
    \end{subfigure}
    ~
    \begin{subfigure}[t]{0.18\linewidth}
        \centering
        \includegraphics[width=\linewidth]{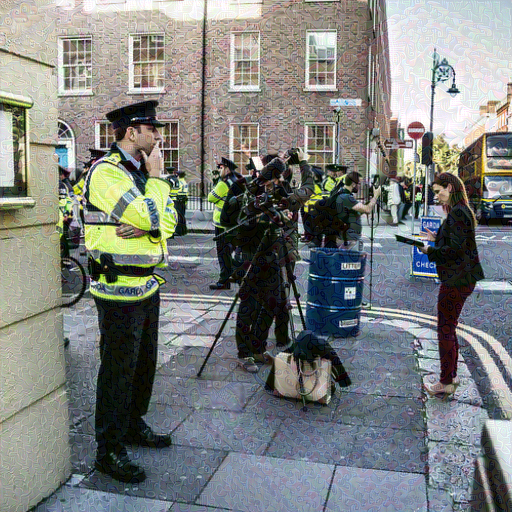}{}
    \end{subfigure}
    ~
    \begin{subfigure}[t]{0.18\linewidth}
        \centering
        \includegraphics[width=\linewidth]{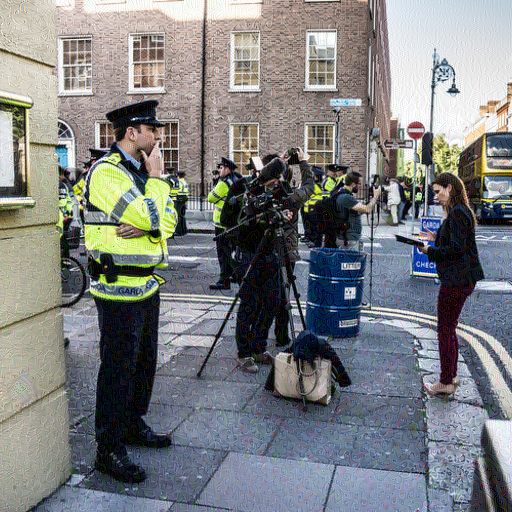}{}
    \end{subfigure}
    ~
    \begin{subfigure}[t]{0.18\linewidth}
        \centering
        \includegraphics[width=\linewidth]{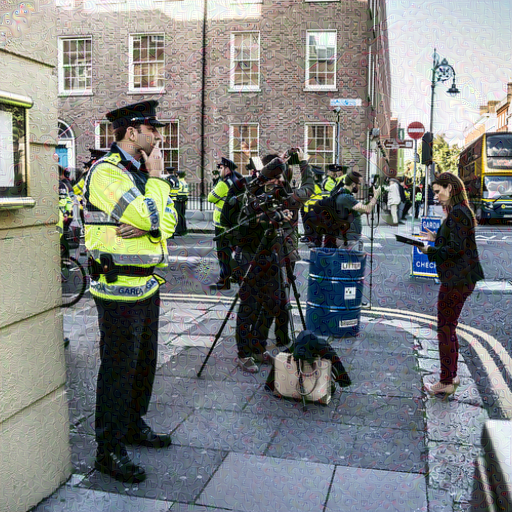}{}
    \end{subfigure}
        \\
         \begin{subfigure}[t]{0.18\linewidth}
        \centering
        \includegraphics[width=\linewidth]{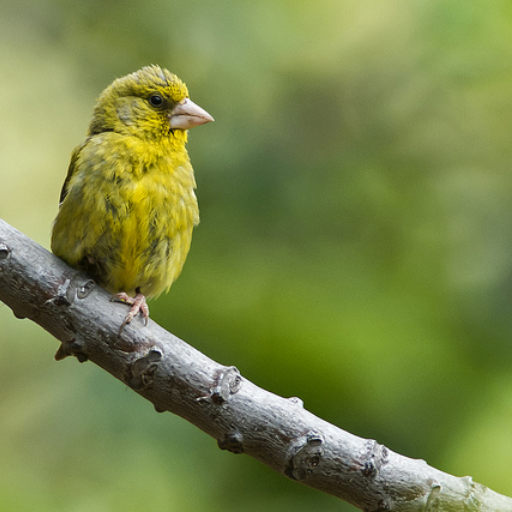}{}
    \end{subfigure}
    ~
     \begin{subfigure}[t]{0.18\linewidth}
        \centering
        \includegraphics[width=\linewidth]{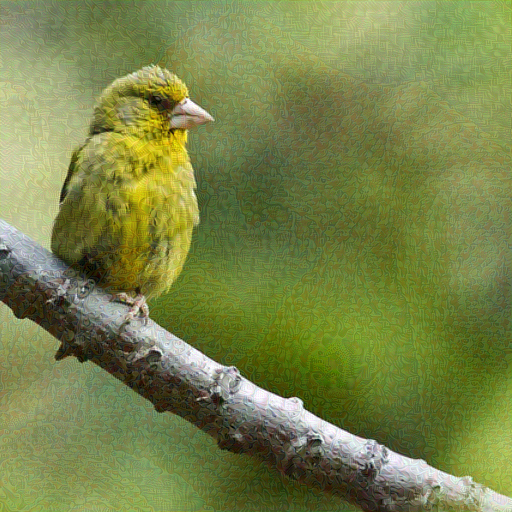}{}
    \end{subfigure}
    ~
    \begin{subfigure}[t]{0.18\linewidth}
        \centering
        \includegraphics[width=\linewidth]{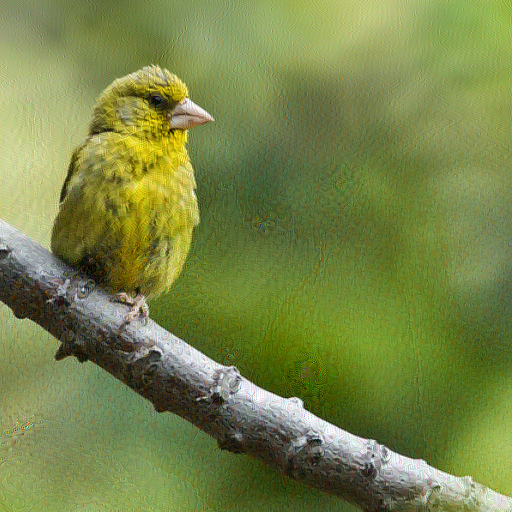}{}
    \end{subfigure}
    ~
    \begin{subfigure}[t]{0.18\linewidth}
        \centering
        \includegraphics[width=\linewidth]{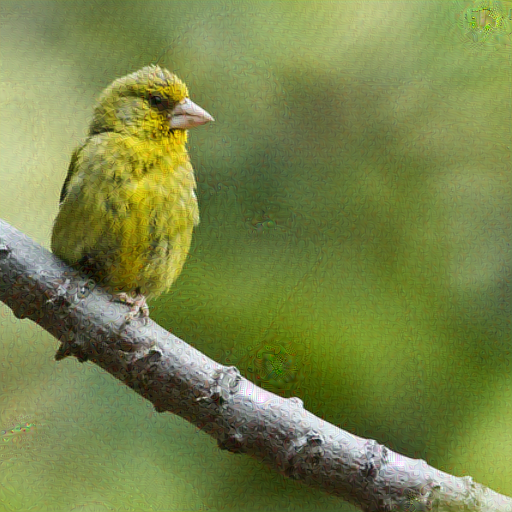}{}
    \end{subfigure}
        \\
    \begin{subfigure}[t]{0.18\linewidth}
        \centering
        \includegraphics[width=\linewidth]{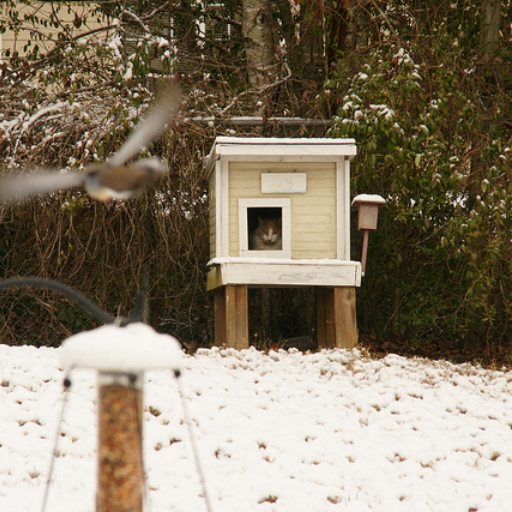}\\
        Original
    \end{subfigure}
    ~
    \begin{subfigure}[t]{0.18\linewidth}
        \centering
        \includegraphics[width=\linewidth]{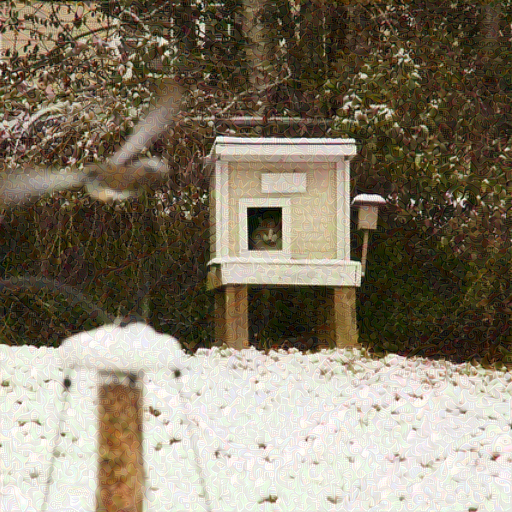}\\
        {Eq. \ref{eq:pgd}}
    \end{subfigure}
    ~
    \begin{subfigure}[t]{0.18\linewidth}
        \centering
        \includegraphics[width=\linewidth]{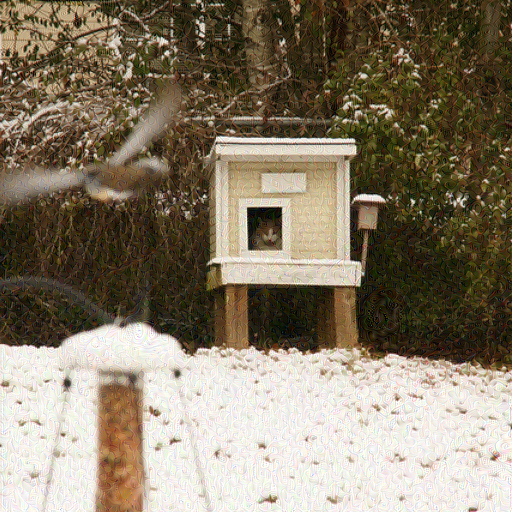}\\
        {Eq. \ref{eq:freq}}
    \end{subfigure}
    ~
    \begin{subfigure}[t]{0.18\linewidth}
        \centering
        \includegraphics[width=\linewidth]{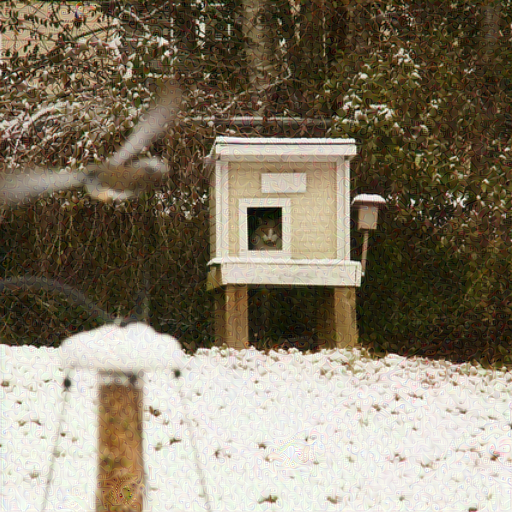}\\
        {Eq. \ref{eq:loss_forgery}}
    \end{subfigure}
    \caption{Comparing trade-off between attack success rate (ASR) and imperceptibility of the attack for different optimization objectives when they achieve similar ASR.}
    \label{fig:ablation}
\end{figure*}

We summarize the results in Table \ref{tab:ablation}, where we show that using the proposed loss formulation in Eq. \ref{eq:loss_forgery} achieves better imperceptibility results with a similar attack success rate. We also show visual examples in Figure \ref{fig:ablation}. Even though the optimization using Eq. \ref{eq:freq} yields lower imperceptibility scores, it can better preserve lower frequency regions, leading to better visual results.

\begin{table}[]
    \centering
    \caption{Comparing trade-off between attack success rate (ASR) and imperceptibility of the attack for different optimization objectives when they achieve similar ASR. We saw that the optimization loss function proposed in Eq. \ref{eq:loss_forgery} does better than others (Eqs. \ref{eq:pgd} and \ref{eq:freq}) at imperceptibility metrics even at higher ASR. We have set hyperparameters $\epsilon$ in Eq. \ref{eq:pgd} to be 0.1, ($\epsilon, \alpha N$) in Eq. \ref{eq:freq} to be (3,300) and $\lambda$ in Eq. \ref{eq:loss_forgery} to be $2.5\times 10^4$. }
    \resizebox{\columnwidth}{!}{
\begin{tabular}{c|ccccccc}
\hline 
Method & ASR & $l_2$ & $l_{\infty}$ & LPIPS & SSIM  & PSNR & FID \\ \hline 
Eq. \ref{eq:pgd} & 83.24 & 68.04 & 0.10 & 0.35 & 0.66 & 28.31 & 69.35 \\
Eq. \ref{eq:freq} & 83.33 & 67.77 & 0.93 & 0.32 & 0.66 & 28.35 & 58.01 \\   
Eq. \ref{eq:loss_forgery} & 84.91 & 44.34 & 0.85 & 0.23 & 0.84 & 31.97 & 56.30 \\
\hline 
\end{tabular}
    }
    \label{tab:ablation}
\end{table}


\section{Watermark Removal using Images with Fixed Pixel Values}
\label{app:remove_127.5}

In Section \ref{sec:attack_forge} for watermark removal, we use images with all pixel values equal to the mean value of the given watermarked image for guidance. In this section, we show that this works better in practice as compared to using a fixed pixel value of 127.5. We report results in Table \ref{tab:127.5}.

\begin{table}[]
    \centering
    \caption{Comparing trade-off between attack success rate (ASR) and imperceptibility of the watermark removal attack when using either images with all pixel value equal to 127.5 or the mean of the watermarked image for guidance. We saw that using the mean value performs better using a fixed pixel value of 127.5. These experiments are run for SDv1.4 for the Gaussian Shading watermarking scheme. We have set the hyperparameter $\lambda$ to be $1\times 10^4$. }
    \resizebox{\columnwidth}{!}{
\begin{tabular}{c|ccccccc}
\hline 
Method & ASR & $l_2$ & $l_{\infty}$ & LPIPS & SSIM  & PSNR & FID \\ \hline 
Mean & 70.10 & 74.10 & 1.31 & 0.29 & 0.77 & 24.12 & 82.75 \\
127.5 & 64.58 &  78.44 &  1.36 &  0.31 &  0.74 &  23.18 &  92.24 \\
\hline 
\end{tabular}
    }
    \label{tab:127.5}
\end{table}

\section{Watermark Removal using Real Images}
\label{app:remove_real}

\begin{figure*}[tbh]
    \centering
    \begin{subfigure}[t]{0.18\linewidth}
        \centering
        \includegraphics[width=\linewidth]{images/108.png}{}
    \end{subfigure}
    ~
    \begin{subfigure}[t]{0.18\linewidth}
        \centering
        \includegraphics[width=\linewidth]{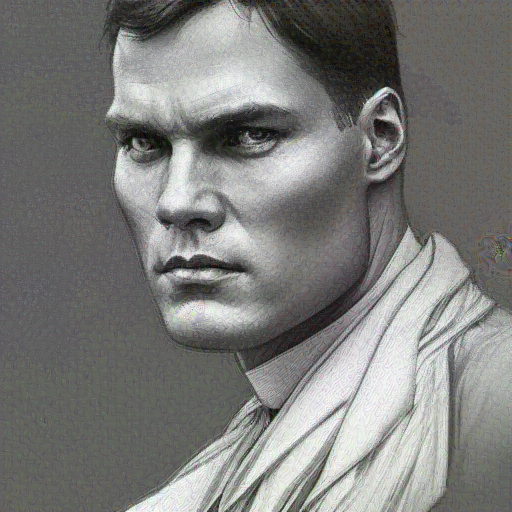}{}
    \end{subfigure}
    ~
    \begin{subfigure}[t]{0.18\linewidth}
        \centering
        \includegraphics[width=\linewidth]{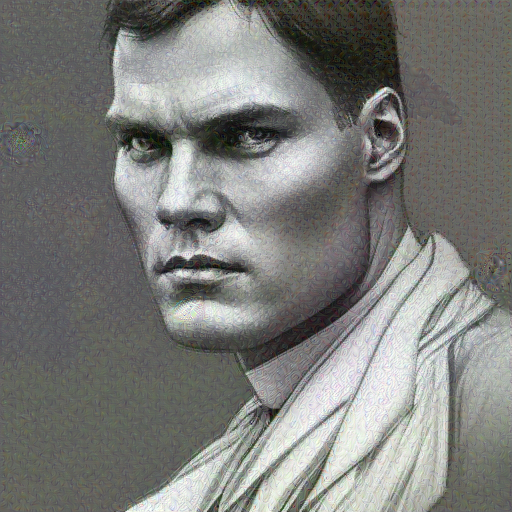}{}
    \end{subfigure}
    ~
    \begin{subfigure}[t]{0.18\linewidth}
        \centering
        \includegraphics[width=\linewidth]{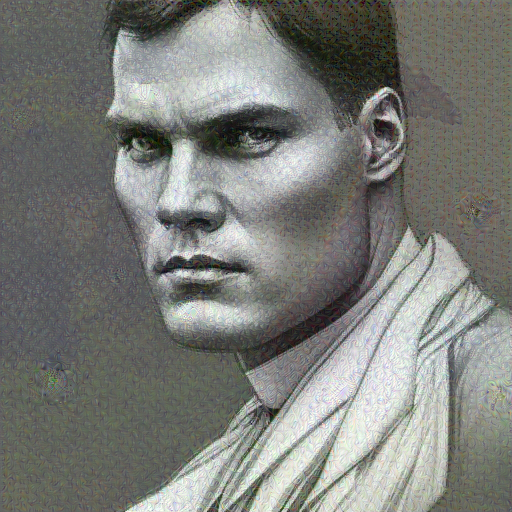}{}
    \end{subfigure}
    \\
    \begin{subfigure}[t]{0.18\linewidth}
        \centering
        \includegraphics[width=\linewidth]{images/11.png}\\
        Original
    \end{subfigure}
    ~
    \begin{subfigure}[t]{0.18\linewidth}
        \centering
        \includegraphics[width=\linewidth]{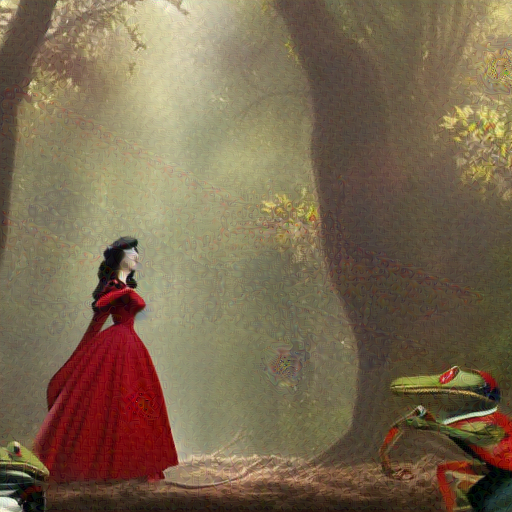}\\
        {$\lambda=5\times 10^4$}
    \end{subfigure}
    ~
    \begin{subfigure}[t]{0.18\linewidth}
        \centering
        \includegraphics[width=\linewidth]{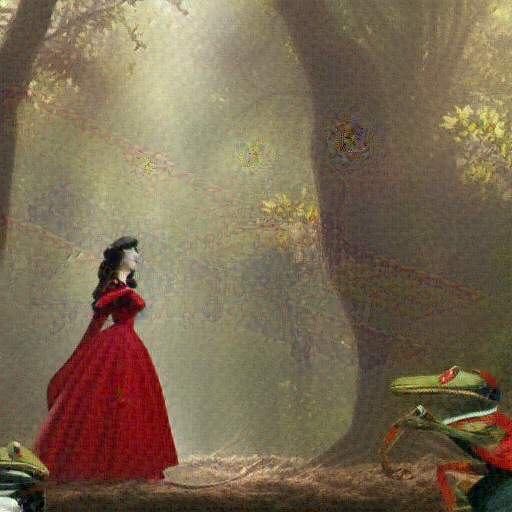}\\
        {$\lambda=2\times 10^4$}
    \end{subfigure}
    ~
    \begin{subfigure}[t]{0.18\linewidth}
        \centering
        \includegraphics[width=\linewidth]{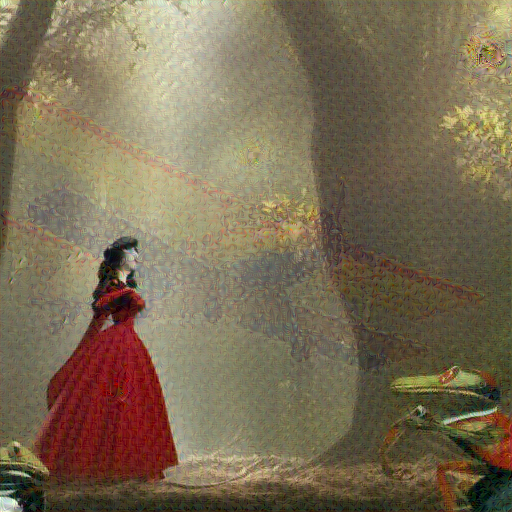}\\
        {$\lambda=1\times 10^4$}
    \end{subfigure}
    \caption{Examples showing successful watermark removal attacks on the Tree-Ring watermarking method with different hyperparameter $\lambda$ values when using \textit{real images for guidance}. }
    \label{fig:tree-rings-examples_removal_appendix_using_real}
\end{figure*}

In this case, we use non-watermarked images from the COCO \cite{lin2014microsoft} dataset for guidance such that we minimize the distance between their respective representations while perturbing the watermarked image. The optimization objective remains the same as Equation \ref{eq:loss_removal}, i.e., 
\begin{equation}
    \min_{\bm{\delta}}  \| \mathcal{E}_{\phi}(\mathbf{x}^{(w)}+ \bm{\delta}) - \mathcal{E}_{\phi}(\mathbf{x}^{(c)}) \|_2 + \lambda \|\bm{\delta}\|_2,
    \label{eq:loss_removal}
\end{equation}
where $\mathbf{x}^{(c)}$ is a randomly selected real image. We show results when using a real image in Table \ref{tab:remove_real_image}. We also show qualitative comparisons in Figure \ref{fig:tree-rings-examples_removal_appendix} and Figure \ref{fig:tree-rings-examples_removal_appendix_using_real}.




\begin{table*}[]
\centering
\caption{Trade-off between attack success rate (ASR) of removing the watermark and imperceptibility of the attack as judged using the $l_2$, $l_{\infty}$ distances and the Learned Perceptual Image Patch Similarity (LPIPS) \cite{zhang2018perceptual}, Structural Similarity Index (SSIM) \cite{wang2004image}, Peak Signal to Noise Ratio (PSNR) and Fréchet Inception Distance (FID) \cite{heusel2017gans} metrics. We use the VAE from SDv1.4 for optimization. The hyperparameter $\lambda$ controls the trade-off between ASR and the amount of perturbation we introduce. }
    \resizebox{0.71\textwidth}{!}{
\begin{tabular}{ccc|ccccccc}
\hline 
Method & Model & $\lambda$ & ASR & $l_2$ & $l_{\infty}$ & LPIPS & SSIM  & PSNR & FID \\ \hline 
\multirow{6}{*}{ RingID \cite{ci2024ringid} } & \multirow{3}{*}{ SDv1.4 } & $5\times 10^4$ & 0.0 & 90.19 & 0.99 & 0.20 & 0.82 & 19.18 & 50.73 \\
& & $2\times 10^4$ & 0.0 & 102.77 & 1.19 & 0.28 & 0.77 & 19.11 & 70.96  \\
& & $1\times 10^4$ & 0.0 & 116.41 & 1.38 & 0.34 & 0.72 & 19.00 & 89.57  \\ \cline{2-10}
\multirow{3}{*}{ } & \multirow{3}{*}{ SDv2.0 } & $5\times 10^4$ & 0.0 & 91.03 & 1.01 & 0.20 & 0.82 & 19.10 & 49.92 \\
& & $2\times 10^4$ & 0.0 & 104.34 & 1.22 & 0.27 & 0.77 & 19.02 & 71.14 \\
& & $1\times 10^4$ & 0.0 & 118.57 & 1.42 & 0.34 & 0.71 & 18.90 & 88.90 \\
\hline 
\multirow{6}{*}{ WIND \cite{arabi2024hidden} } & \multirow{3}{*}{ SDv1.4 } & $5\times 10^4$ & 0.0 & 124.82 & 1.18 & 0.24 & 0.78 & 17.47 & 65.91  \\
& & $2\times 10^4$ & 0.0 & 134.89 & 1.31 & 0.31 & 0.73 & 17.45 & 84.33  \\
& & $1\times 10^4$ & 0.0 & 146.23 & 1.44 & 0.37 & 0.68 & 17.40 & 103.29   \\ \cline{2-10}
&  \multirow{3}{*}{ SDv2.0 } & $5\times 10^4$ & 0.0 & 95.29 & 1.03 & 0.20 & 0.82 & 18.73 & 54.46 \\
& & $2\times 10^4$ & 0.0 & 108.22 & 1.24 & 0.27 & 0.77 & 18.67 & 76.59 \\
& & $1\times 10^4$ & 0.0 & 122.17 & 1.43 & 0.34 & 0.71 & 18.56 & 95.83 \\
\hline 
\end{tabular}
}
\label{tab:remove_wind_ringid}
\end{table*}

\begin{table*}[]
\centering
\caption{Trade-off between attack success rate (ASR) of removing the watermark and imperceptibility of the attack as judged using the $l_2$, $l_{\infty}$ distances and the Learned Perceptual Image Patch Similarity (LPIPS) \cite{zhang2018perceptual}, Structural Similarity Index (SSIM) \cite{wang2004image}, Peak Signal to Noise Ratio (PSNR) and Fréchet Inception Distance (FID) \cite{heusel2017gans} metrics. We use the VAE from SDv1.4 for optimization. The hyperparameter $\lambda$ controls the trade-off between ASR and the amount of perturbation we introduce. }
    \resizebox{0.71\textwidth}{!}{
\begin{tabular}{ccc|ccccccc}
\hline 
Method & Model & $\lambda$ & ASR & $l_2$ & $l_{\infty}$ & LPIPS & SSIM  & PSNR & FID \\ \hline 
\multirow{6}{*}{ Tree-Ring \cite{wen2023tree} } & \multirow{3}{*}{ SDv1.4 } & $5\times 10^4$ & 68.23 & 80.67 & 0.86 & 0.25 & 0.82 & 20.61 & 66.94  \\ 
& & $2\times 10^4$ & 78.94 & 98.55 & 1.03 & 0.35 & 0.74 & 20.39 & 94.32 \\
& & $1\times 10^4$ & 83.62 & 117.55 & 1.15 & 0.45 & 0.66 & 20.05 & 121.82  \\
\cline{2-10}
\multirow{3}{*}{ } & \multirow{3}{*}{ SDv2.0 } & $5\times 10^4$ & 61.62 & 91.47 & 0.98 & 0.27 & 0.80 & 19.85 & 73.39  \\
& & $2\times 10^4$ & 70.81 & 108.30 & 1.12 & 0.37 & 0.73 & 19.65 & 97.78  \\
& & $1\times 10^4$ & 74.59 & 126.83 & 1.22 & 0.46 & 0.65 & 19.37 & 125.82 \\
\hline 
\multirow{6}{*}{ RingID \cite{ci2024ringid} } & \multirow{3}{*}{ SDv1.4 } & $5\times 10^4$ & 0.0 & 90.02 & 0.84 & 0.24 & 0.81 & 19.34 & 60.87   \\
& & $2\times 10^4$ & 1.10 & 106.56 & 1.03 & 0.33 & 0.74 & 19.21 & 83.20   \\
& & $1\times 10^4$ & 1.65 & 125.16 & 1.19 & 0.42 & 0.67 & 18.98 & 108.07  \\ \cline{2-10}
\multirow{3}{*}{ } & \multirow{3}{*}{ SDv2.0 } & $5\times 10^4$ & 0.0 & 91.14 & 0.83 & 0.24 & 0.81 & 19.23 & 58.94 \\
& & $2\times 10^4$ & 0.54 & 107.50 & 1.05 & 0.33 & 0.74 & 19.15 & 82.27 \\
& & $1\times 10^4$ & 0.54 & 126.90 & 1.21 & 0.41 & 0.67 & 18.90 & 104.30 \\
\hline 
\multirow{6}{*}{ WIND \cite{arabi2024hidden} } & \multirow{3}{*}{ SDv1.4 } & $5\times 10^4$ & 0.0 & 126.55 & 1.11 & 0.28 & 0.77 & 17.50 & 75.41 \\
& & $2\times 10^4$ & 1.09 & 139.97 & 1.19 & 0.36 & 0.70 & 17.44 & 99.79 \\
& & $1\times 10^4$ & 1.09 & 155.78 & 1.32 & 0.44 & 0.63 & 17.31 & 121.98 \\ \cline{2-10}
&  \multirow{3}{*}{ SDv2.0 } & $5\times 10^4$ & 0.0 & 94.25 & 0.85 & 0.24 & 0.81 & 18.94 & 63.91 \\
& & $2\times 10^4$ & 0.54 & 111.23 & 1.05 & 0.33 & 0.74 & 18.81 & 89.86 \\
& & $1\times 10^4$ & 0.54 & 130.19 & 1.20 & 0.41 & 0.67 & 18.59 & 112.61 \\
\hline 
\multirow{6}{*}{ Gaussian Shading \cite{yang2024gaussian}} & \multirow{3}{*}{ SDv1.4 } & $5\times 10^4$ & 28.0 & 46.65 & 0.84 & 0.21 & 0.84 & 28.36 & 64.71  \\
& & $2\times 10^4$ & 65.0 & 73.49 & 1.08 & 0.32 & 0.75 & 23.87 & 93.28 \\
& & $1\times 10^4$ & 81.0 & 92.20 & 1.21 & 0.42 & 0.67 & 23.70 & 118.36  \\ 
\cline{2-10}
&  \multirow{3}{*}{ SDv2.0 } & $5\times 10^4$ & 25.0 & 44.31 & 0.84 & 0.18 & 0.86 & 31.58 & 53.57 \\
& & $2\times 10^4$ & 49.0 & 65.25 & 1.06 & 0.29 & 0.78 & 28.44 & 80.45 \\
& & $1\times 10^4$ & 74.0 & 87.61 & 1.20 & 0.39 & 0.70 & 25.96 & 104.90  \\ 
\hline 
\end{tabular}
}
\label{tab:remove_real_image}
\end{table*}

\section{Detailed Experiment Settings}
\label{app:implementation}

\subsection{Prompt Datasets}

We utilize two datasets namely,  the \textit{Gustavosta/Stable-Diffusion-Prompts}\footnote{\href{https://huggingface.co/datasets/Gustavosta/Stable-Diffusion-Prompts}{https://huggingface.co/datasets/Gustavosta/Stable-Diffusion-Prompts}} dataset and the \textit{runwayml-stable-diffusion-v1-5-eval-random-prompts}\footnote{\href{https://huggingface.co/datasets/yuvalkirstain/runwayml-stable-diffusion-v1-5-eval-random-prompts}{https://huggingface.co/datasets/yuvalkirstain/runwayml-stable-diffusion-v1-5-eval-random-prompts}} dataset. The former contains around 80,000 prompts extracted from the image finder for Stable Diffusion: "Lexica.art". These are generally longer and more detailed prompts. The latter, on the other hand, contains 200 short and simple prompts that contain less information/ details. 

\subsection{Watermarking Methods}

We consider four latent-noise based watermarking schemes - Tree-Rings \cite{wen2023tree}, RingID \cite{ci2024ringid}, Gaussian Shading \cite{yang2024gaussian} and WIND \cite{arabi2024hidden}. We utilized their publicly available implementations from the following GitHub repositories, 
\begin{itemize}
    \item Tree-Ring: \href{https://github.com/YuxinWenRick/tree-ring-watermark}{https://github.com/YuxinWenRick/tree-ring-watermark}.
    \item RingID: \href{https://github.com/showlab/RingID}{https://github.com/showlab/RingID}.
    \item Gaussian Shading \href{https://github.com/bsmhmmlf/Gaussian-Shading/}{https://github.com/bsmhmmlf/Gaussian-Shading/}.
    \item WIND \href{https://github.com/anonymousiclr2025submission/Hidden-in-the-Noise}{https://github.com/anonymousiclr2025submission/Hidden-in-the-Noise}.
\end{itemize}

\subsection{Hyperparameters}

We utilize the following hyperparameters in our optimization:
\begin{itemize}
    \item Number of iterations: 15,000
    \item Learning rate $\alpha$: 0.020
    \item Image size: $512 \times 512$
    \item Stable Diffusion versions: CompVis/stable-diffusion-v1-4 and stabilityai/stable-diffusion-2
    
\end{itemize}

\section{More Visual Examples}
\label{app:visual}

\begin{figure*}[tbh]
    \centering
    \begin{subfigure}[t]{0.18\linewidth}
        \centering
        \includegraphics[width=\linewidth]{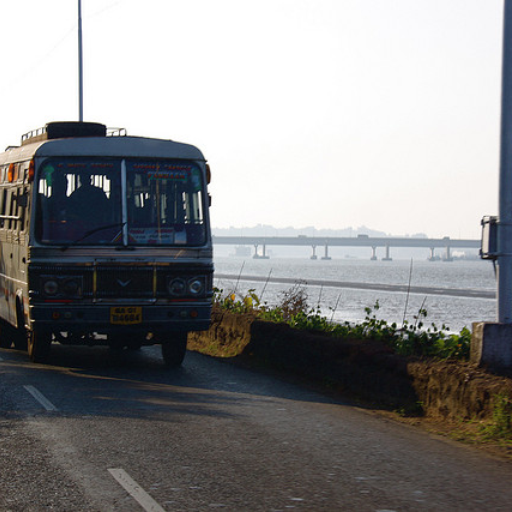}{}
    \end{subfigure}
    ~
    \begin{subfigure}[t]{0.18\linewidth}
        \centering
        \includegraphics[width=\linewidth]{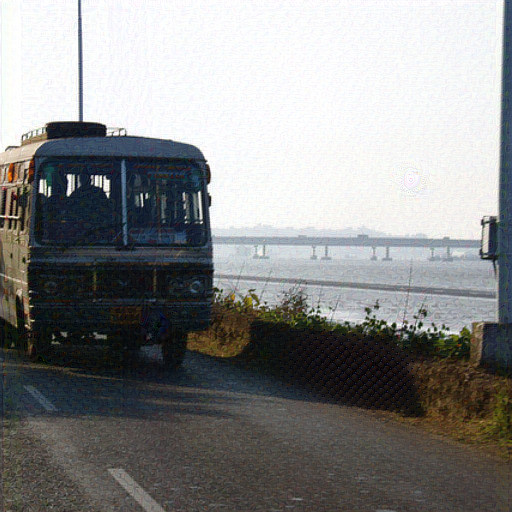}{}
    \end{subfigure}
    ~
    \begin{subfigure}[t]{0.18\linewidth}
        \centering
        \includegraphics[width=\linewidth]{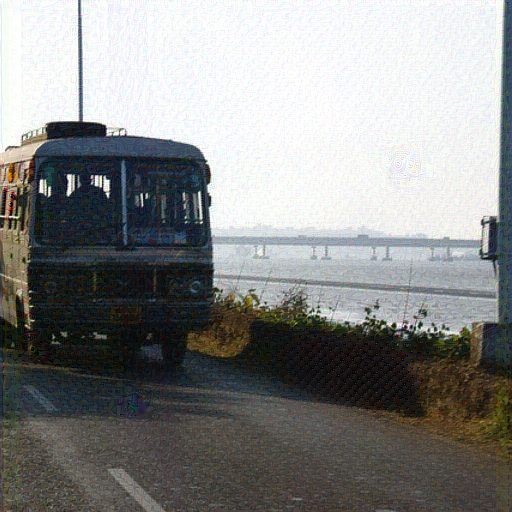}{}
    \end{subfigure}
    ~
    \begin{subfigure}[t]{0.18\linewidth}
        \centering
        \includegraphics[width=\linewidth]{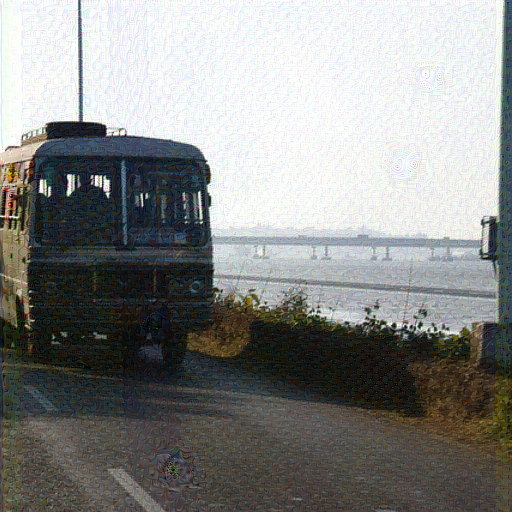}{}
    \end{subfigure}
        \\
    \begin{subfigure}[t]{0.18\linewidth}
        \centering
        \includegraphics[width=\linewidth]{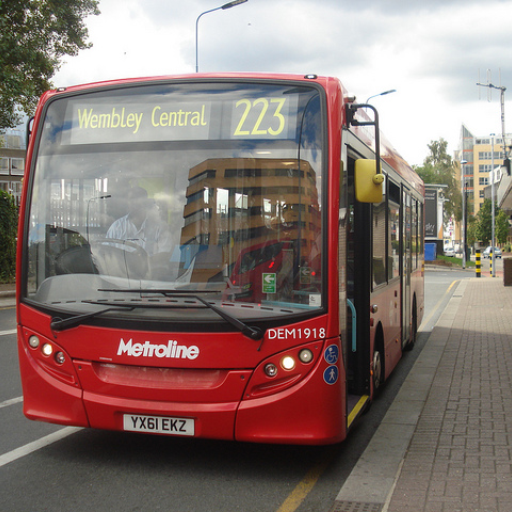}\\
        Original
    \end{subfigure}
    ~
    \begin{subfigure}[t]{0.18\linewidth}
        \centering
        \includegraphics[width=\linewidth]{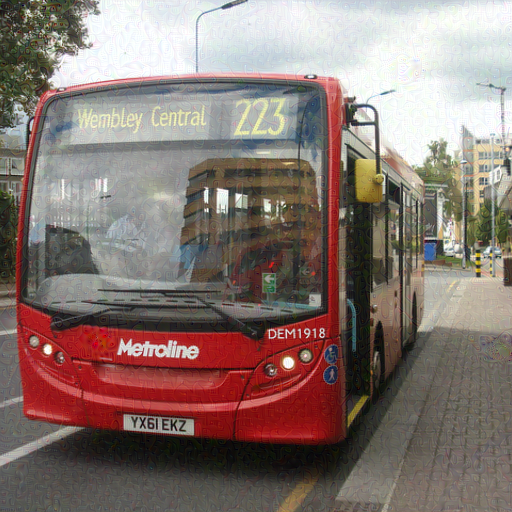}\\
        {$\lambda=5\times 10^4$}
    \end{subfigure}
    ~
    \begin{subfigure}[t]{0.18\linewidth}
        \centering
        \includegraphics[width=\linewidth]{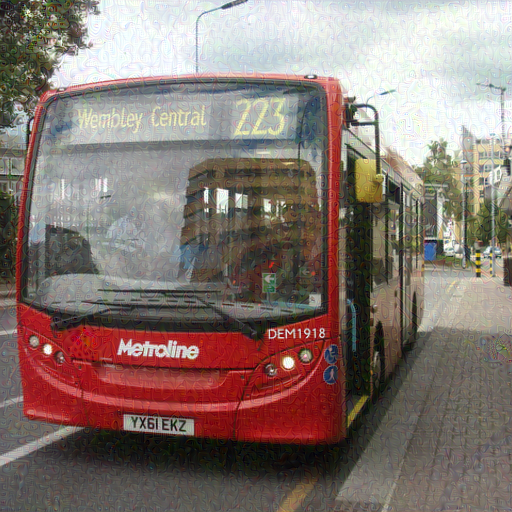}\\
        {$\lambda=2\times 10^4$}
    \end{subfigure}
    ~
    \begin{subfigure}[t]{0.18\linewidth}
        \centering
        \includegraphics[width=\linewidth]{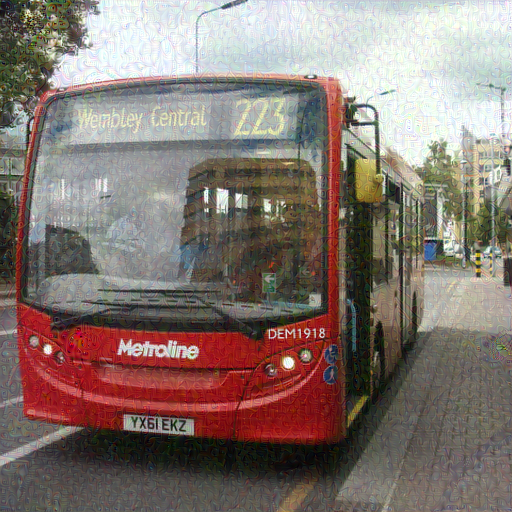}\\
        {$\lambda=1\times 10^4$}
    \end{subfigure}
    \caption{Examples showing successful watermark forgery attacks on the Tree-Ring watermarking method with different hyperparameter $\lambda$ values. }
    \label{fig:tree-rings-examples_appendix}
\end{figure*}

\begin{figure*}[tbh]
    \centering
    \begin{subfigure}[t]{0.18\linewidth}
        \centering
        \includegraphics[width=\linewidth]{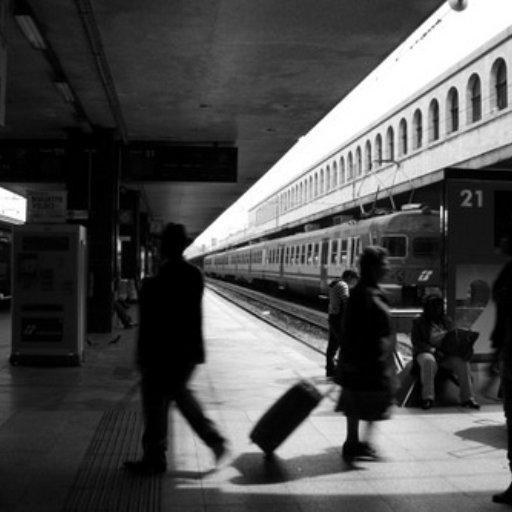}{}
    \end{subfigure}
    ~
    \begin{subfigure}[t]{0.18\linewidth}
        \centering
        \includegraphics[width=\linewidth]{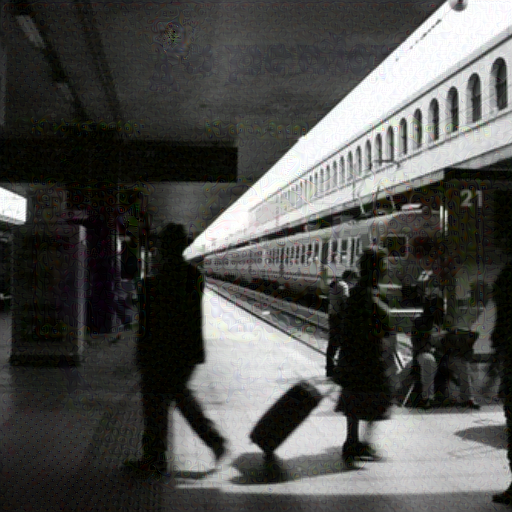}{}
    \end{subfigure}
    ~
    \begin{subfigure}[t]{0.18\linewidth}
        \centering
        \includegraphics[width=\linewidth]{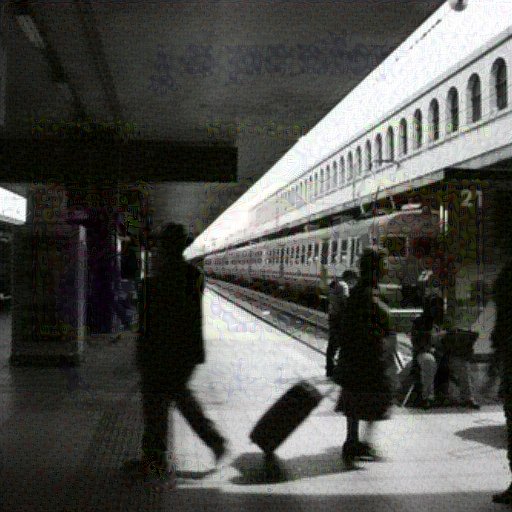}{}
    \end{subfigure}
    ~
    \begin{subfigure}[t]{0.18\linewidth}
        \centering
        \includegraphics[width=\linewidth]{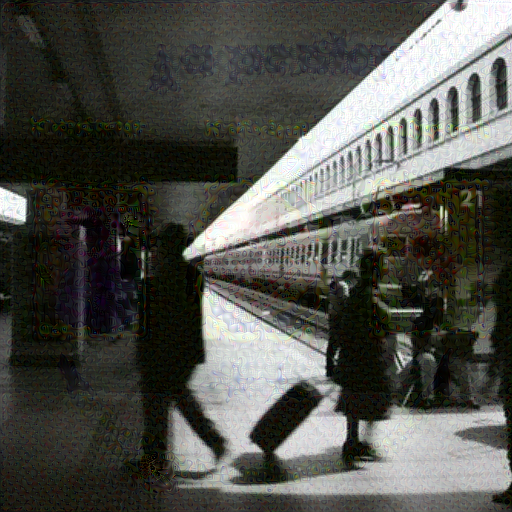}{}
    \end{subfigure}
        \\
    \begin{subfigure}[t]{0.18\linewidth}
        \centering
        \includegraphics[width=\linewidth]{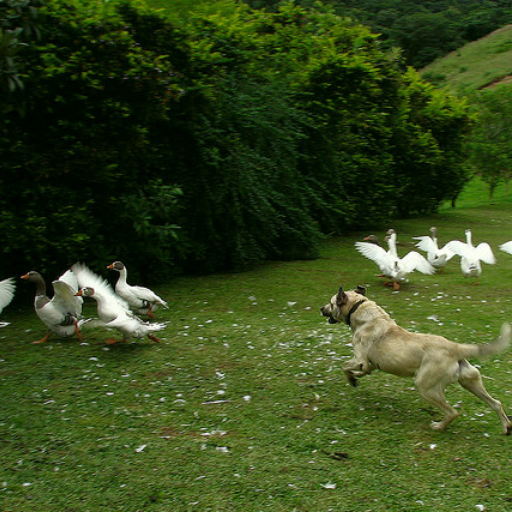}\\
        Original
    \end{subfigure}
    ~
    \begin{subfigure}[t]{0.18\linewidth}
        \centering
        \includegraphics[width=\linewidth]{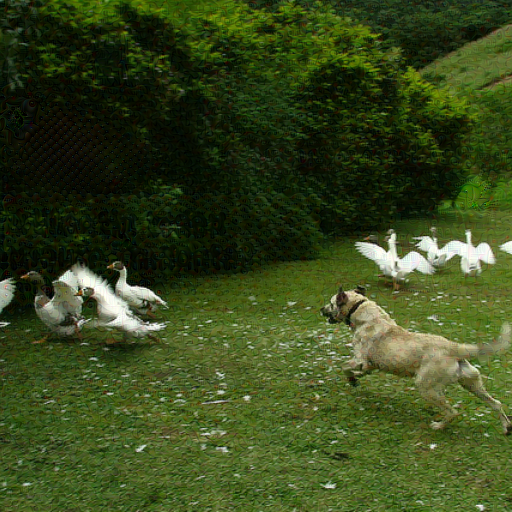}\\
        {$\lambda=5\times 10^4$}
    \end{subfigure}
    ~
    \begin{subfigure}[t]{0.18\linewidth}
        \centering
        \includegraphics[width=\linewidth]{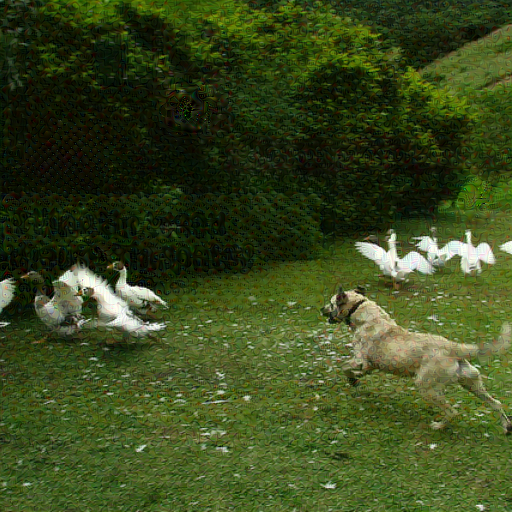}\\
        {$\lambda=2\times 10^4$}
    \end{subfigure}
    ~
    \begin{subfigure}[t]{0.18\linewidth}
        \centering
        \includegraphics[width=\linewidth]{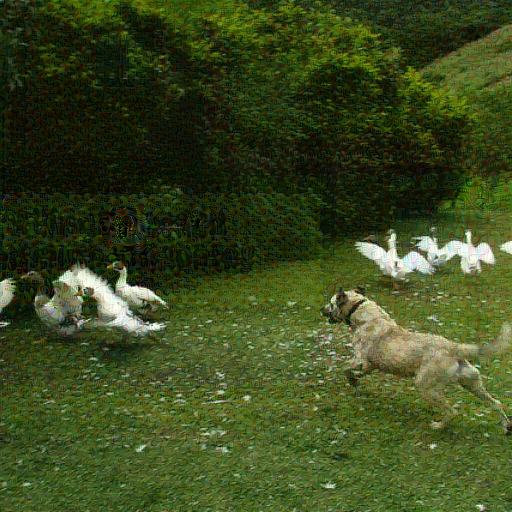}\\
        {$\lambda=1\times 10^4$}
    \end{subfigure}
    \caption{Examples showing successful watermark forgery attacks on the RingID watermarking method with different hyperparameter $\lambda$ values. }
    \label{fig:ringid_appendix}
\end{figure*}

\begin{figure*}[tbh]
    \centering
    \begin{subfigure}[t]{0.18\linewidth}
        \centering
        \includegraphics[width=\linewidth]{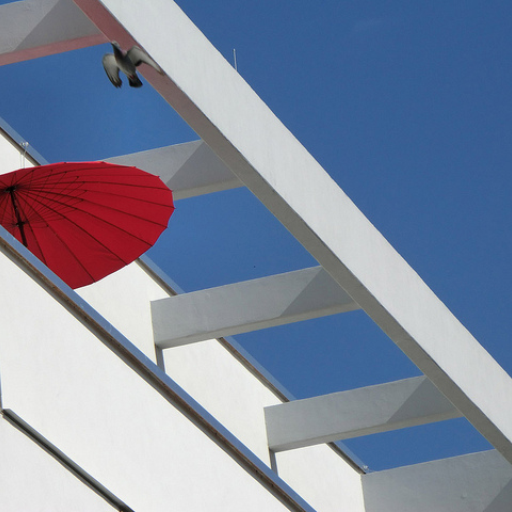}{}
    \end{subfigure}
    ~
    \begin{subfigure}[t]{0.18\linewidth}
        \centering
        \includegraphics[width=\linewidth]{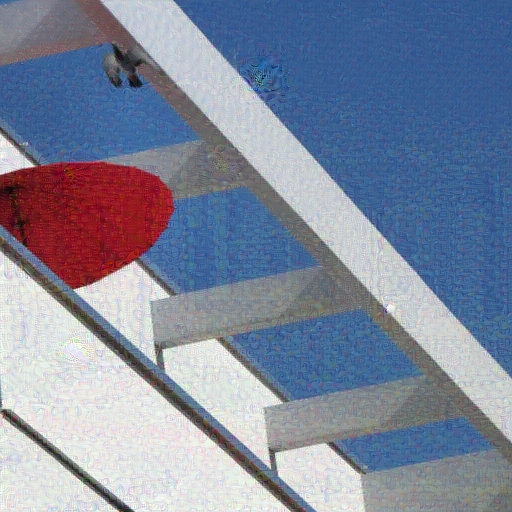}{}
    \end{subfigure}
    ~
    \begin{subfigure}[t]{0.18\linewidth}
        \centering
        \includegraphics[width=\linewidth]{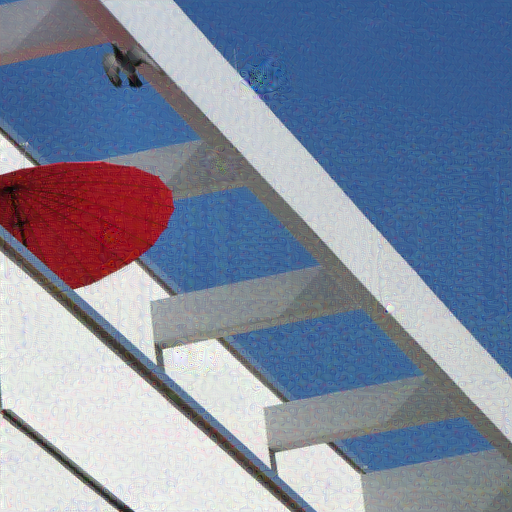}{}
    \end{subfigure}
    ~
    \begin{subfigure}[t]{0.18\linewidth}
        \centering
        \includegraphics[width=\linewidth]{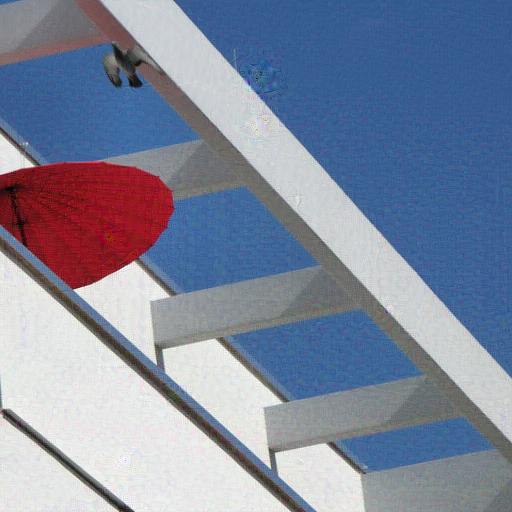}{}
    \end{subfigure}
        \\
    \begin{subfigure}[t]{0.18\linewidth}
        \centering
        \includegraphics[width=\linewidth]{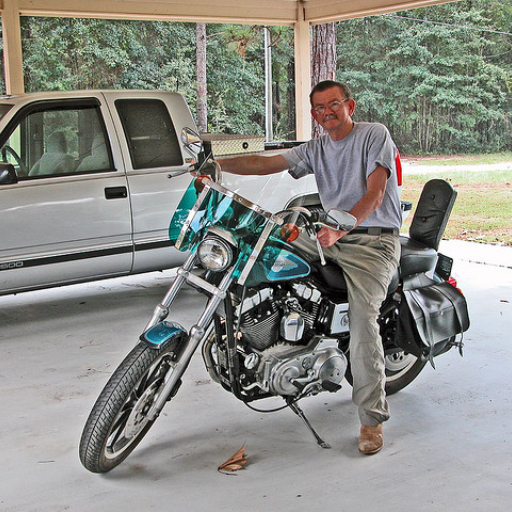}\\
        Original
    \end{subfigure}
    ~
    \begin{subfigure}[t]{0.18\linewidth}
        \centering
        \includegraphics[width=\linewidth]{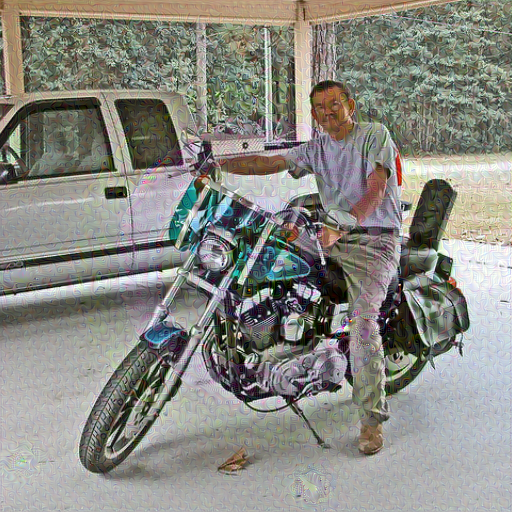}\\
        {$\lambda=5\times 10^4$}
    \end{subfigure}
    ~
    \begin{subfigure}[t]{0.18\linewidth}
        \centering
        \includegraphics[width=\linewidth]{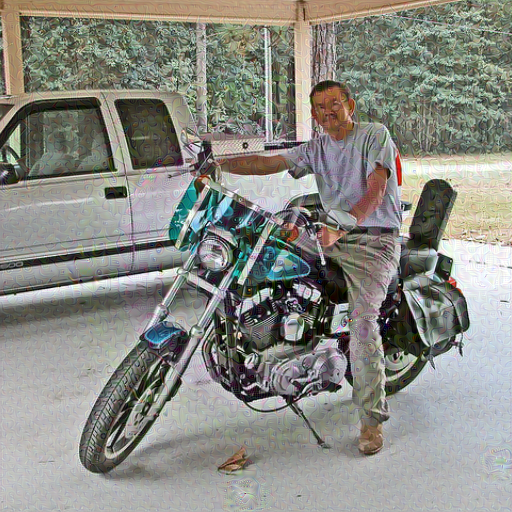}\\
        {$\lambda=2\times 10^4$}
    \end{subfigure}
    ~
    \begin{subfigure}[t]{0.18\linewidth}
        \centering
        \includegraphics[width=\linewidth]{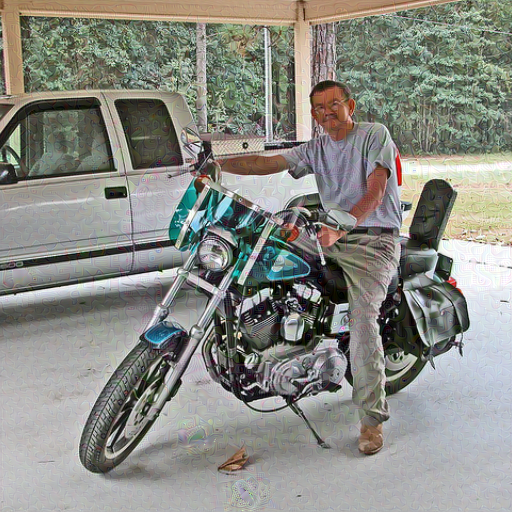}\\
        {$\lambda=1\times 10^4$}
    \end{subfigure}
    \caption{Examples showing successful watermark forgery attacks on the WIND watermarking method with different hyperparameter $\lambda$ values. }
    \label{fig:wind_appendix}
\end{figure*}

\begin{figure*}[tbh]
    \centering
    \begin{subfigure}[t]{0.18\linewidth}
        \centering
        \includegraphics[width=\linewidth]{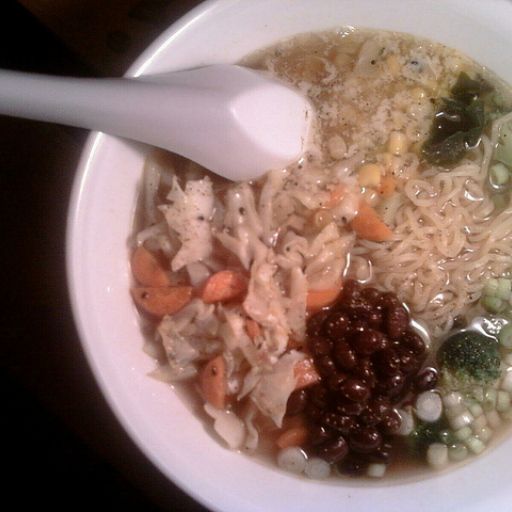}{}
    \end{subfigure}
    ~
    \begin{subfigure}[t]{0.18\linewidth}
        \centering
        \includegraphics[width=\linewidth]{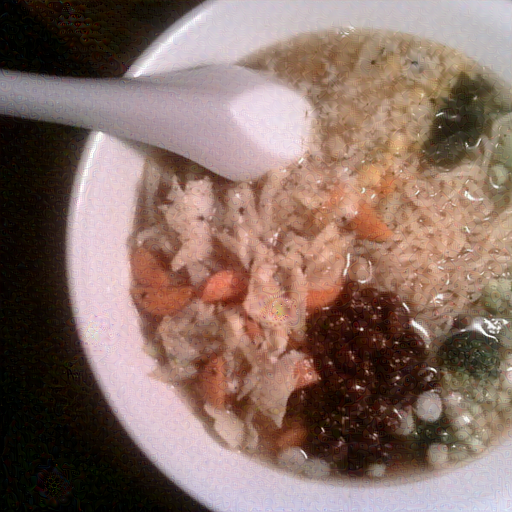}{}
    \end{subfigure}
    ~
    \begin{subfigure}[t]{0.18\linewidth}
        \centering
        \includegraphics[width=\linewidth]{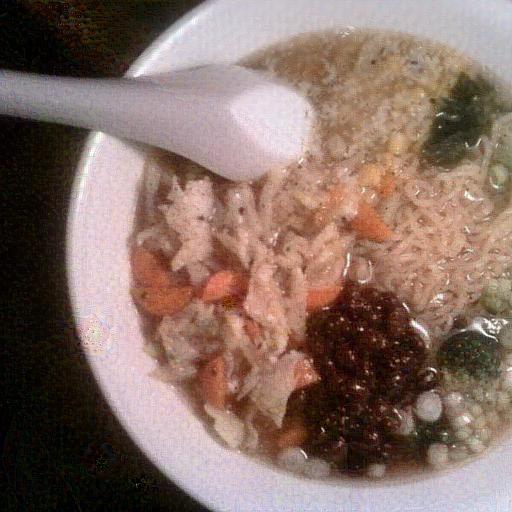}{}
    \end{subfigure}
    ~
    \begin{subfigure}[t]{0.18\linewidth}
        \centering
        \includegraphics[width=\linewidth]{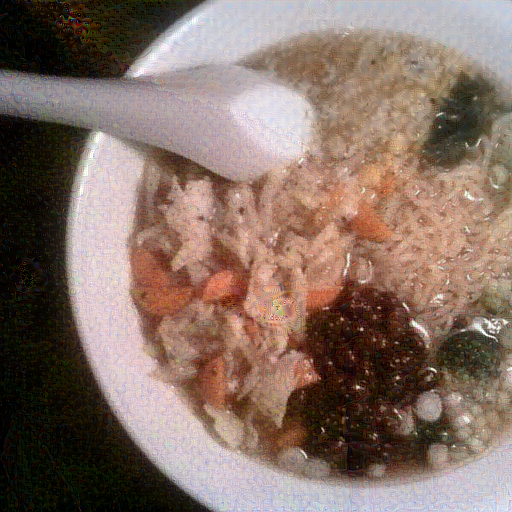}{}
    \end{subfigure}
        \\
    \begin{subfigure}[t]{0.18\linewidth}
        \centering
        \includegraphics[width=\linewidth]{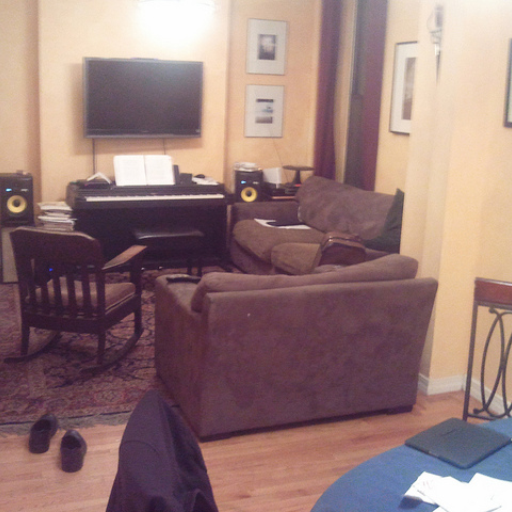}\\
        Original
    \end{subfigure}
    ~
    \begin{subfigure}[t]{0.18\linewidth}
        \centering
        \includegraphics[width=\linewidth]{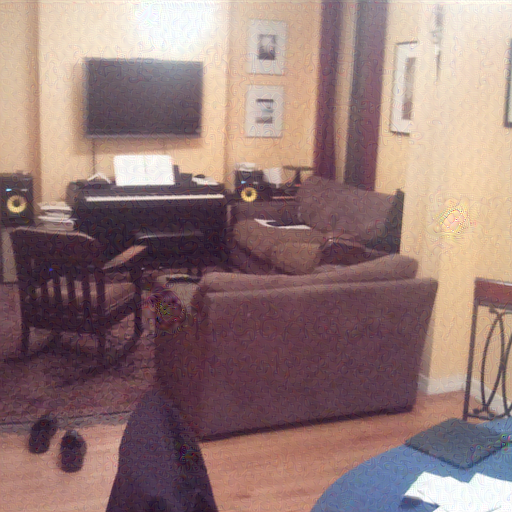}\\
        {$\lambda=5\times 10^4$}
    \end{subfigure}
    ~
    \begin{subfigure}[t]{0.18\linewidth}
        \centering
        \includegraphics[width=\linewidth]{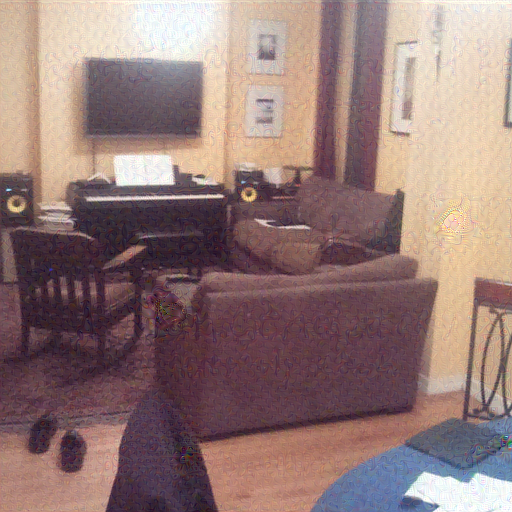}\\
        {$\lambda=2\times 10^4$}
    \end{subfigure}
    ~
    \begin{subfigure}[t]{0.18\linewidth}
        \centering
        \includegraphics[width=\linewidth]{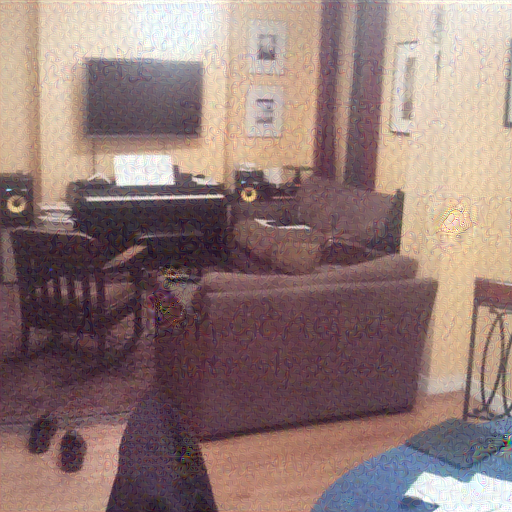}\\
        {$\lambda=1\times 10^4$}
    \end{subfigure}
    \caption{Examples showing successful watermark forgery attacks on the Gaussian Shading watermarking method with different hyperparameter $\lambda$ values. }
    \label{fig:gaussian_shading_appendix}
\end{figure*}

\begin{figure*}[tbh]
    \centering
    \begin{subfigure}[t]{0.18\linewidth}
        \centering
        \includegraphics[width=\linewidth]{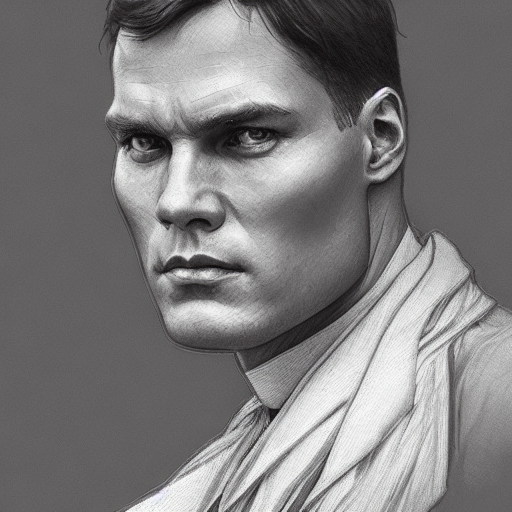}{}
    \end{subfigure}
    ~
    \begin{subfigure}[t]{0.18\linewidth}
        \centering
        \includegraphics[width=\linewidth]{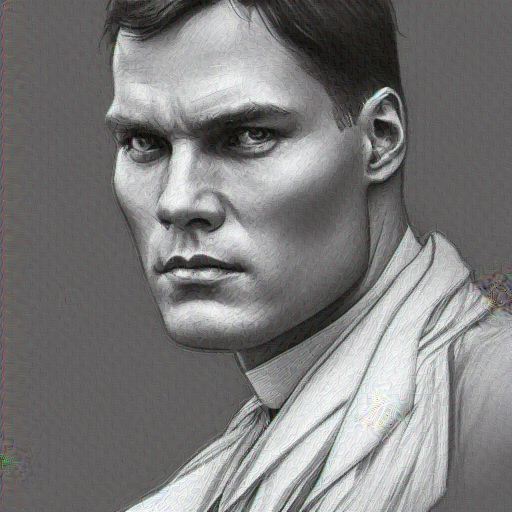}{}
    \end{subfigure}
    ~
    \begin{subfigure}[t]{0.18\linewidth}
        \centering
        \includegraphics[width=\linewidth]{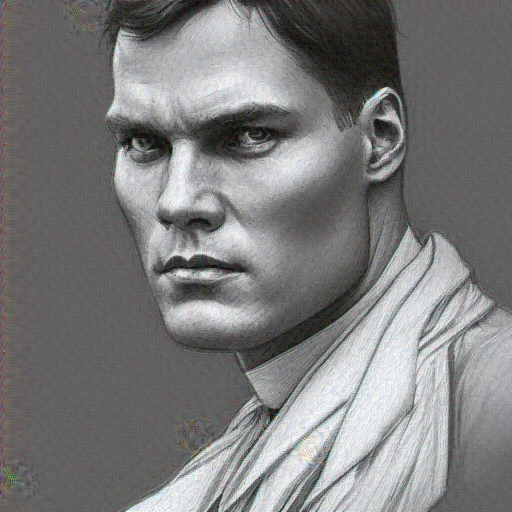}{}
    \end{subfigure}
    ~
    \begin{subfigure}[t]{0.18\linewidth}
        \centering
        \includegraphics[width=\linewidth]{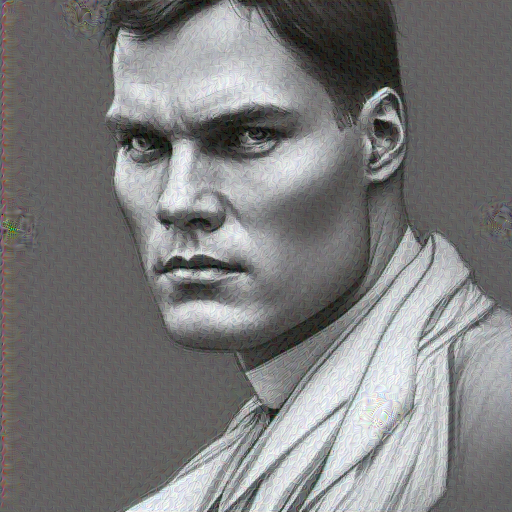}{}
    \end{subfigure}
    \\
    \begin{subfigure}[t]{0.18\linewidth}
        \centering
        \includegraphics[width=\linewidth]{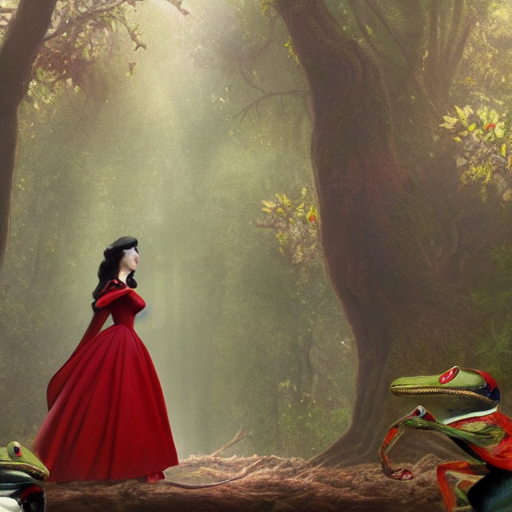}\\
        Original
    \end{subfigure}
    ~
    \begin{subfigure}[t]{0.18\linewidth}
        \centering
        \includegraphics[width=\linewidth]{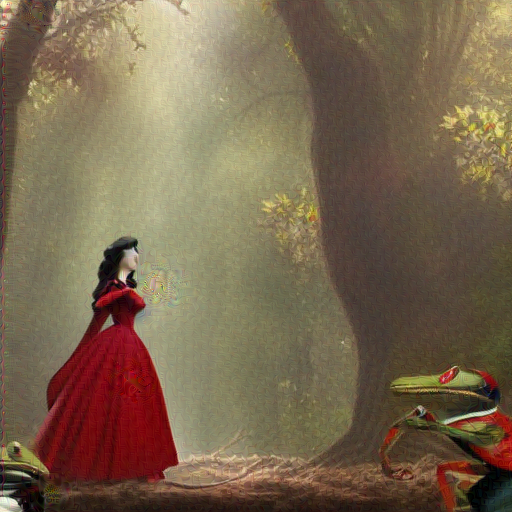}\\
        {$\lambda=5\times 10^4$}
    \end{subfigure}
    ~
    \begin{subfigure}[t]{0.18\linewidth}
        \centering
        \includegraphics[width=\linewidth]{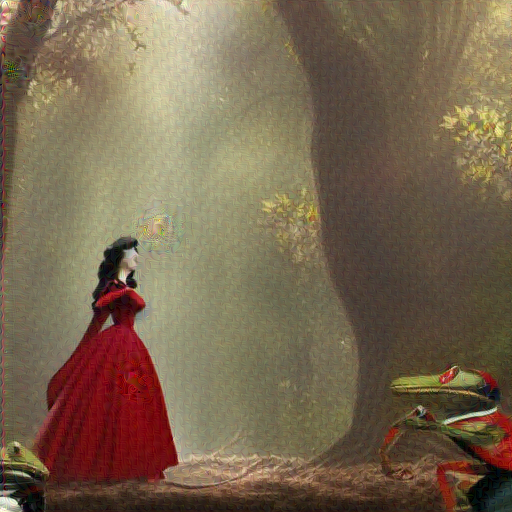}\\
        {$\lambda=2\times 10^4$}
    \end{subfigure}
    ~
    \begin{subfigure}[t]{0.18\linewidth}
        \centering
        \includegraphics[width=\linewidth]{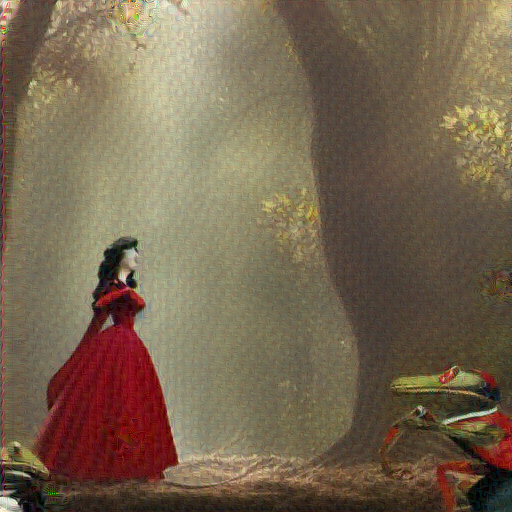}\\
        {$\lambda=1\times 10^4$}
    \end{subfigure}
    \caption{Examples showing successful watermark removal attacks on the Tree-Ring watermarking method with different hyperparameter $\lambda$ values when using \textit{an image with all pixels equal to the mean of the watermarked image for guidance}. }
    \label{fig:tree-rings-examples_removal_appendix}
\end{figure*}

\begin{figure*}[tbh]
    \centering
    \begin{subfigure}[t]{0.18\linewidth}
        \centering
        \includegraphics[width=\linewidth]{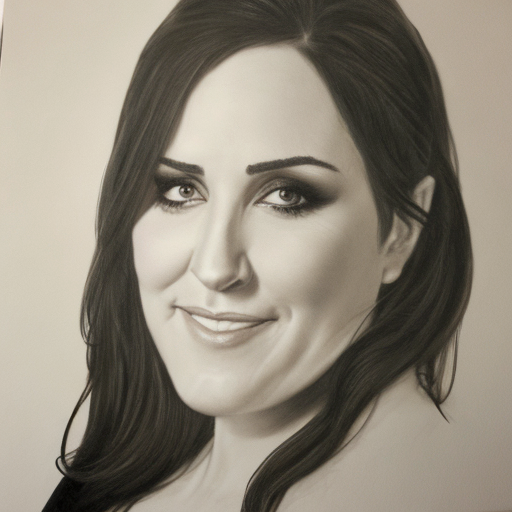}{}
    \end{subfigure}
    ~
    \begin{subfigure}[t]{0.18\linewidth}
        \centering
        \includegraphics[width=\linewidth]{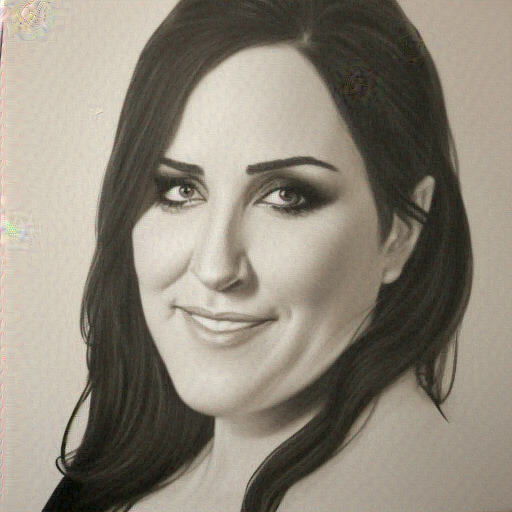}{}
    \end{subfigure}
    ~
    \begin{subfigure}[t]{0.18\linewidth}
        \centering
        \includegraphics[width=\linewidth]{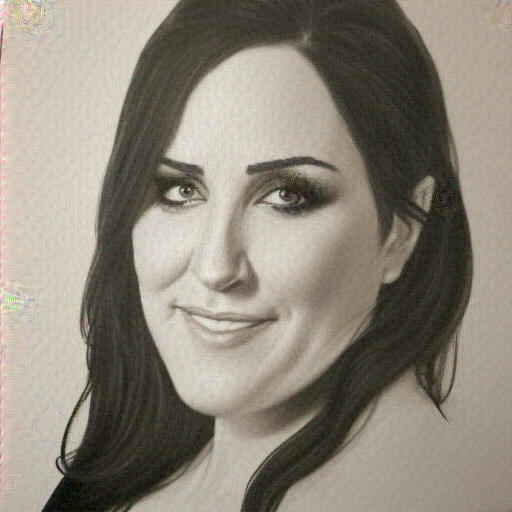}{}
    \end{subfigure}
    ~
    \begin{subfigure}[t]{0.18\linewidth}
        \centering
        \includegraphics[width=\linewidth]{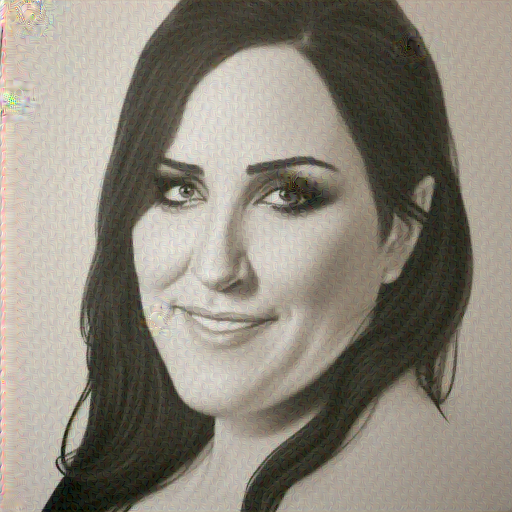}{}
    \end{subfigure}
    \\
    \begin{subfigure}[t]{0.18\linewidth}
        \centering
        \includegraphics[width=\linewidth]{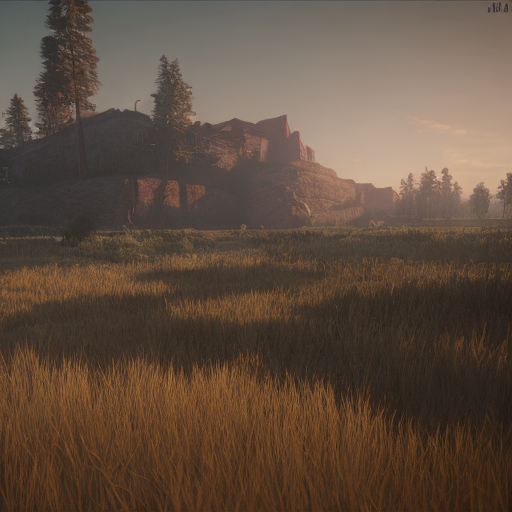}\\
        Original
    \end{subfigure}
    ~
    \begin{subfigure}[t]{0.18\linewidth}
        \centering
        \includegraphics[width=\linewidth]{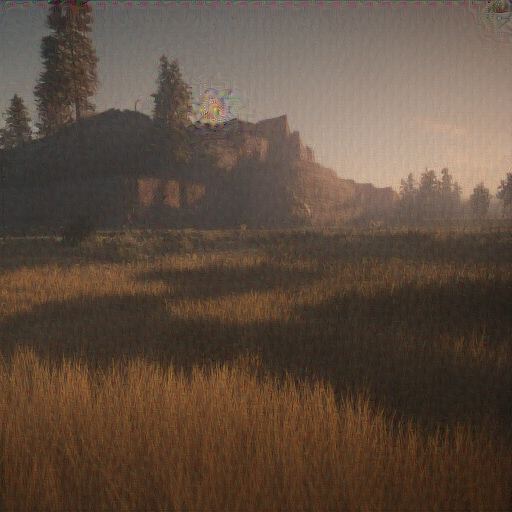}\\
        {$\lambda=5\times 10^4$}
    \end{subfigure}
    ~
    \begin{subfigure}[t]{0.18\linewidth}
        \centering
        \includegraphics[width=\linewidth]{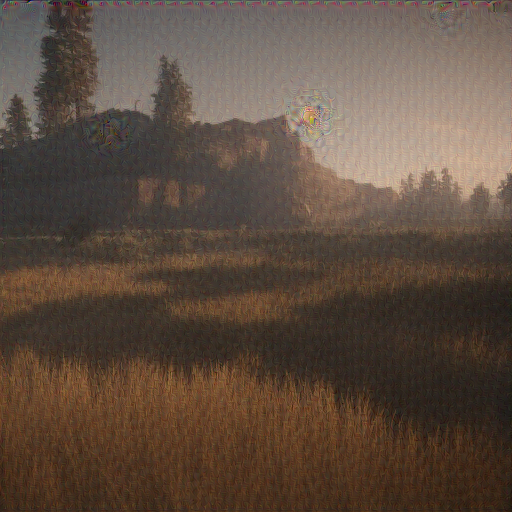}\\
        {$\lambda=2\times 10^4$}
    \end{subfigure}
    ~
    \begin{subfigure}[t]{0.18\linewidth}
        \centering
        \includegraphics[width=\linewidth]{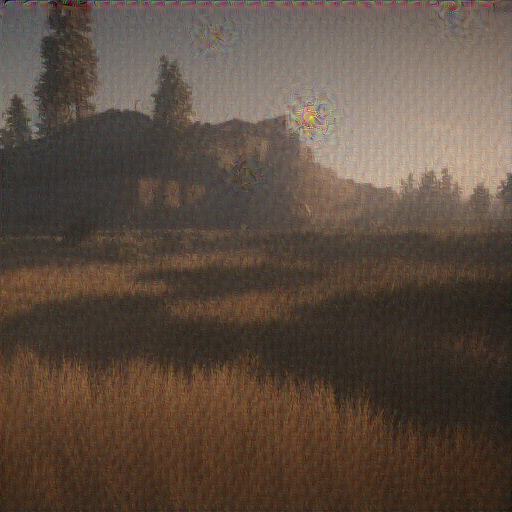}\\
        {$\lambda=1\times 10^4$}
    \end{subfigure}
    \caption{Examples showing successful watermark removal attacks on the Gaussian Shading watermarking method with hyperparameter $\lambda$ values. }
    \label{fig:gaussian_removal}
\end{figure*}

We provide additional visual examples to show that successful watermark forgery and removal attacks do not harm the semantics or quality of the resultant image. We show examples of forgery attacks on Tree-Ring in Figure \ref{fig:tree-rings-examples_appendix}, RingID in Figure \ref{fig:ringid_appendix}, WIND in Figure \ref{fig:wind_appendix}, and Gaussian Shading in Figure \ref{fig:gaussian_shading_appendix}. We show examples of the removal attack in Figure \ref{fig:tree-rings-examples_removal_appendix} for the Tree-Ring system and in Figure \ref{fig:gaussian_removal} for the Gaussian Shading system.

\end{document}